\documentclass[sigconf,screen,svgnames,table,hyphens,hidelinks]{acmart}

\setcopyright{none}

\acmConference[ISSTA '20]{The 29th ACM SIGSOFT International Symposium
  on Software Testing and Analysis}{18--22 July 2020}{Los Angeles,
  California, United States}

\usepackage{xspace}
\usepackage{booktabs}   
\usepackage{subcaption} 
\usepackage{xcolor}
\usepackage{textcomp}
\usepackage{pgfpages}
\usepackage{pgfplots}
\usepackage{pgfplotstable}
\usepackage{tikz}
\usepackage{enumerate}
\usepackage{algorithmicx}
\usepackage{algorithm}
\usepackage[noend]{algpseudocode}
\usepackage{multirow}
\usepackage{amsmath}
\usepackage{amsthm}
\usepackage{listings}
\usepackage{colortbl}
\usepackage{url}
\usepackage{titlecaps}
\usepackage[font=bf]{caption}
\usepackage{flushend}
\usepackage{booktabs}
\usepackage{colortbl}
\usepackage{graphicx}
\usepackage{siunitx}

\definecolor{gray}{rgb}{0.5, 0.5, 0.5}
\definecolor{light-gray}{gray}{0.77}
\definecolor{BrickRed}{rgb}{0.8, 0.25, 0.33}
\definecolor{Black}{rgb}{0.0, 0.0, 0.0}
\definecolor{DarkBlue}{rgb}{0.0, 0.0, 0.55}
\definecolor{Crimson}{rgb}{0.86, 0.08, 0.24}
\definecolor{SlateGrey}{rgb}{0.44, 0.5, 0.56}
\definecolor{lightorange}{HTML}{FFB74D}
\definecolor{blue}{rgb}{0.0, 0.0, 1.0}
\definecolor{magenta}{rgb}{0.79, 0.08, 0.48}

\makeatletter
\newenvironment{btHighlight}[1][]
{\begingroup\tikzset{bt@Highlight@par/.style={#1}}\begin{lrbox}{\@tempboxa}}
{\end{lrbox}\bt@HL@box[bt@Highlight@par]{\@tempboxa}\endgroup}

\newcommand\btHL[1][]{%
  \begin{btHighlight}[#1]\bgroup\aftergroup\bt@HL@endenv%
}
\def\bt@HL@endenv{%
  \end{btHighlight}%
  \egroup
}
\newcommand{\bt@HL@box}[2][]{%
  \tikz[#1]{%
    \pgfpathrectangle{\pgfpoint{0.3pt}{0pt}}{\pgfpoint{\wd #2}{\ht #2}}%
    \pgfusepath{use as bounding box}%
    \node[anchor=base west,fill=lightorange,outer sep=0pt,inner xsep=0.3pt,inner ysep=0pt,minimum height=\ht\strutbox+0.3pt,#1]{\raisebox{0.3pt}{\strut}\strut\usebox{#2}};
  }%
}
\makeatother

\makeatletter
\newif\if@anonymize

\@anonymizetrue

\if@anonymize
  \newcommand\anonymize[1]{Link removed for double-blind review.}
\else
  \newcommand\anonymize[1]{\tiny#1}
\fi
\makeatother

\usetikzlibrary{calc,trees,positioning,arrows,chains,shapes.geometric,%
  decorations.pathreplacing,decorations.pathmorphing,decorations.text,shapes,%
  matrix,shapes.symbols,patterns,shadows,automata}

\def\addlegendimage{\csname pgfplots@addlegendimage\endcsname}

\pgfplotsset{compat=newest}

\algrenewcommand\alglinenumber[1]{\tiny\color{Black!70}{#1}}
\algrenewcommand\algorithmicforall[2]{\textbf{for} $i=$ #1 \textbf{to} #2}
\algrenewtext{ForAll}{\algorithmicforall}
\algnewcommand\algorithmicswitch{\textbf{switch}}
\algnewcommand\algorithmiccase{\textbf{case}}
\algdef{SE}[SWITCH]{Switch}{EndSwitch}[1]{\algorithmicswitch\ #1\ \algorithmicdo}{\algorithmicend\ \algorithmicswitch}%
\algdef{SE}[CASE]{Case}{EndCase}[1]{\algorithmiccase\ #1}{\algorithmicend\ \algorithmiccase}%
\algtext*{EndSwitch}%
\algtext*{EndCase}%
\algdef{SE}[SUBALG]{Indent}{EndIndent}{}{\algorithmicend\ }%
\algtext*{Indent}
\algtext*{EndIndent}

\lstdefinestyle{basic}{%
  morekeywords     = [1]{var},%
  morekeywords     = [2]{},%
  keywordstyle     = [2]\color{teal}\bfseries,%
  morekeywords     = [3]{minimize},%
  keywordstyle     = [3]\color{BrickRed}\bfseries,%
  keywordstyle     = \bfseries\color{DarkBlue},%
  commentstyle     = \ttfamily\color{Black!70}\lst@ifdisplaystyle\footnotesize\fi,%
  basicstyle       = \ttfamily\lst@ifdisplaystyle\footnotesize\fi,%
  emph             = {int,char,double,float,unsigned,void,bool},%
  emphstyle        = {\color{teal}\bfseries},%
  columns          = [c]fixed,%
  aboveskip        = 0mm,%
  belowskip        = 2mm,%
  keepspaces       = true,%
  mathescape       = true,%
  escapechar       = ¤,%
  tabsize          = 2,%
  numbers          = left,%
  numberstyle      = \tiny\color{Black!70},%
  numbersep        = 4pt,%
  stepnumber       = 1,%
  firstnumber      = 1,%
  showstringspaces = false,%
  captionpos       = b,%
  extendedchars    = true,%
  upquote          = true,%
  abovecaptionskip = 0mm,%
  belowcaptionskip = 0mm,%
  moredelim        = **[is][{\btHL[fill=light-gray]}]{°}{°},%
}

\lstdefinestyle{clang}{%
  language         = C,%
  style            = basic,%
}

\newcommand\secref[1]{Sect.~\ref{#1}}
\newcommand\figref[1]{Fig.~\ref{#1}}
\newcommand\tabref[1]{Tab.~\ref{#1}}
\newcommand\figsref[1]{Figs.~\ref{#1}}
\newcommand\tabsref[1]{Tabs.~\ref{#1}}
\newcommand\tool{\textsc{RAID}\xspace}
\begin{document}

\title[\tool: Randomized Adversarial-Input Detection for Neural Networks]{\tool: Randomized Adversarial-Input Detection\\for Neural Networks}

\author{Hasan Ferit Eniser}
\affiliation{
  \institution{MPI-SWS \country{Germany}}
}
\email{hfeniser@mpi-sws.org}

\author{Maria Christakis}
\affiliation{
  \institution{MPI-SWS \country{Germany}}
}
\email{maria@mpi-sws.org}

\author{Valentin W{\"u}stholz}
\affiliation{
  \institution{ConsenSys/MythX \country{Germany}}
}
\email{valentin.wustholz@consensys.net}

\begin{abstract}
  In recent years, neural networks have become the default choice for image classification
  and many other learning tasks, even though they are vulnerable to so-called
  adversarial attacks. To increase their robustness against these attacks, there have
  emerged numerous detection mechanisms that aim to automatically determine if an
  input is adversarial. However, state-of-the-art detection mechanisms either rely on
  being tuned for each type of attack, or they do not generalize across different attack
  types. To alleviate these issues, we propose a novel technique for adversarial-image
  detection, \tool, that trains a secondary classifier to identify differences in neuron
  activation values between benign and adversarial inputs. Our technique is both more
  reliable and more effective than the state of the art when evaluated against six popular
  attacks. Moreover, a straightforward extension of \tool increases its robustness against
  detection-aware adversaries without affecting its effectiveness.
\end{abstract}

\maketitle





\section{Introduction}
\label{sect:intro}

There is no doubt that neural networks are becoming rapidly and
increasingly prevalent. Their success has been particularly impressive
for the task of accurately recognizing patterns and classifying
images~\cite{SzegedyVanhoucke2016}, on which we focus here. Even
though such networks are able to achieve very high accuracy for
``normal'' (i.e., benign) images, they may be tricked by adversaries
into providing wrong classifications. More specifically, given a
correctly classified image, an adversary may perturb it
slightly---\emph{typically almost unnoticeably according to human
  perception}---to generate an image that is classified
differently. Such images are referred to as
\emph{adversarial}~\cite{SzegedyZaremba2014} and pose a serious threat
to emerging applications of machine learning, such as autonomous
driving~\cite{PapernotMcDaniel2017,EykholtEvtimov2018}.

To protect neural networks against adversarial attacks, there have
emerged numerous \emph{defense} mechanisms that aim to correctly
classify adversarial inputs. However, most of these defenses have not
been found effective in preventing adversarial images from being
misclassified~\cite{CarliniAthalye2019}.
As a result, the research community has also focused on automated
\emph{detection} of adversarial images, that is, on devising
mechanisms for detecting whether an input to a neural network is
adversarial (see \cite{CarliniWagner2017-Bypassing} for examples).

\paragraph{\textbf{Adversarial-image detection.}}
In the context of adversarial-image detection, some state-of-the-art
detection mechanisms are classifier based, that is, they train a
secondary classifier for determining whether inputs of a neural
network are adversarial. A recent, notable example is
SADL~\cite{KimFeldt2019}, an approach that relies on the observation
that ``surprising'' inputs are more likely to be
adversarial. Depending on the surprise measure, the effectiveness of
this approach relies on tuning its hyper-parameters, namely which
neurons are used to measure surprise, for each type of
attack. However, in practice, detection mechanisms are typically unaware of the
types of incoming attacks.

In terms of classifier-free detection mechanisms, a novel example is
mMutant~\cite{WangDong2019}, which assumes adversarial images to be
more ``sensitive'' than normal ones. More specifically, the assumption
is that the classification outcome for adversarial inputs is more
likely to change when performing minor mutations to the neural
network. However, this assumption does not generalize to adversarial
inputs with a high prediction confidence, that is, inputs for which
the neural network provides a wrong classification with high
confidence.

\paragraph{\textbf{Our approach.}}
To alleviate these issues, we present a new adversarial-image
detection technique, called \tool. Our technique is based on teaching
a secondary classifier to recognize differences in neuron activation
values between adversarial and normal inputs. We show that the
effectiveness of \tool is stable with respect to its hyper-parameters
across a wide range of adversarial attacks. \tool consistently
outperforms SADL and mMutant on these attacks by up to 88\%. Moreover,
in contrast to these techniques, there exists a simple extension to
\tool that increases its robustness against detection-aware
adversaries without affecting its effectiveness.

\paragraph{\textbf{Contributions.}} Our paper makes the following
contributions:
\begin{enumerate}
\item We present a simple, yet very effective, adversarial-image
  detection technique for neural networks.
\item We extend our technique to increase its robustness against
  stronger adversaries that can tailor their attacks to specific
  detection mechanisms.
\item We implement our approach and make the implementation publicly
  available.
\item We extensively evaluate our approach and compare it with three
  state-of-the-art detection techniques.
\end{enumerate}

\paragraph{\textbf{Outline.}}
The next section provides background on different types of adversaries and
attacks. \secref{sect:approach} explains our adversarial-image detection approach. We
present our experimental evaluation in \secref{sect:experiments}, discuss related work in
\secref{sect:relatedWork}, and conclude in \secref{sect:conclusion}.

\section{Background}
\label{sect:background}
In this section, we give a short overview of threat models and
adversarial attacks.

\subsection{Threat Models}
\label{subsect:models}

A \emph{threat model} describes the conditions under which a detection
mechanism is designed to work. Consequently, the threat model is
necessary for assessing the effectiveness of a
detector~\cite{CarliniAthalye2019}.

Adversarial attacks are typically categorized according to two main threat
models: (1)~\textit{white-box} attacks, where the adversary has
perfect knowledge of the neural network including, for example, its
architecture and parameters, and (2)~\textit{black-box} attacks, which
generate adversarial examples without any internal information about
the neural network. In this work, we consider the stronger white-box
threat model although our technique is also applicable against
black-box attacks. White-box adversaries come in two capacities,
namely static and adaptive.

\paragraph{\textbf{Static adversaries.}}
A \emph{static adversary} is an attacker that is unaware of any
detection mechanism protecting a network model against adversarial
attacks. A static adversary makes use of existing white-box attacks to
generate adversarial examples but does not tailor these attacks to
breach any specific detection mechanism.

\paragraph{\textbf{Adaptive adversaries.}}
An \emph{adaptive adversary} is an attacker that is aware of the
detection mechanism protecting a network model, if any. Such an
adversary also has knowledge of internal parameters of the detection
mechanism. As a result, it may adapt the adversarial attacks it
generates to breach the particular detection mechanism that is in
place. Adaptive adversaries are clearly more powerful than static
ones.\smallskip

In this paper, we evaluate our approach as well as three
state-of-the-art adversarial-image detection techniques on several
static adversaries. We also extend our approach to adaptive
adversaries and reason about its effectiveness.

\subsection{Adversarial Attacks}
\label{subsect:attacks}

In the context of deep learning, adversarial examples are generally
defined as inputs to a neural network that are specifically crafted to
trick it into making a wrong prediction~\cite{CarliniAthalye2019}.
Practically, such examples are generated by slightly distorting
correctly classified inputs.
Since discovering the vulnerability of neural networks to adversarial
examples~\cite{SzegedyZaremba2014}, there have emerged numerous
attacks in the literature, some of which have gained significant
traction and are routinely used as benchmarks for evaluating both
other attacks as well as detection mechanisms.

In our experiments, we also use such well-known attacks to evaluate
the effectiveness of our approach against static adversaries. More
specifically, we use the following six attacks:
\begin{enumerate}
\item Projected Gradient Descent (\textbf{PGD})~\cite{MadryMakelov2018}
\item Fast Gradient Sign Method (\textbf{FGSM})~\cite{GoodfellowShlens2015}
\item Basic Iterative Method (\textbf{BIM})~\cite{KurakinGoodfellow2017}
\item DeepFool (\textbf{DF})~\cite{MoosaviDezfooliFawzi2016}
\item Carlini-Wagner (\textbf{CW})~\cite{CarliniWagner2017-Robustness}
\item Jacobian Saliency Map Attack
  (\textbf{JSMA})~\cite{PapernotMcDaniel2016-Limitations}
\end{enumerate}

Each of these attacks specifies an upper bound on the amount of
allowed distortion. This bound is typically defined in terms of norms,
such as $\text{L}_\infty$, $\text{L}_2$, and $\text{L}_0$. For
example, DF is an $\text{L}_2$ attack, meaning that the $\text{L}_2$
norm of original and adversarial inputs cannot be larger than a given
bound. Regarding the above attacks, PGD, FGSM, and BIM are
$\text{L}_\infty$ attacks, DF and CW $\text{L}_2$, and JSMA
$\text{L}_0$. The $\text{L}_2$ attacks are found to be stronger than
the others~\cite{CarliniWagner2017-Robustness}.

As we previously mentioned, these are white-box attacks of static
adversaries. Attacks of adaptive adversaries are specifically tailored
to the detection mechanism of the model under threat. Therefore,
different detection mechanisms require different attacks, which means
that there is no off-the-shelf adaptive adversary.

\section{Our Approach}
\label{sect:approach}

\begin{figure}[t]
	\includegraphics[scale=0.25]{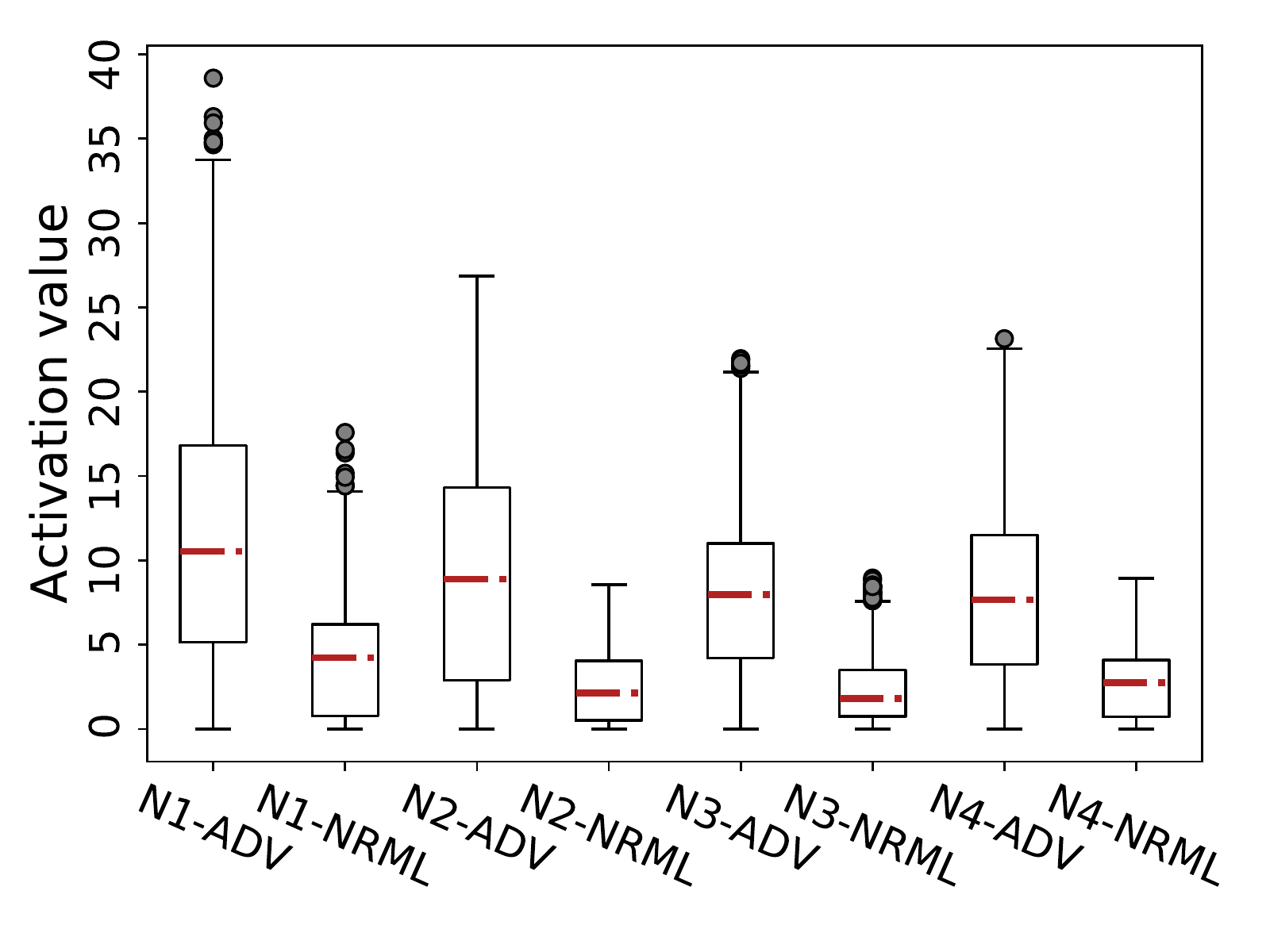}
	\includegraphics[scale=0.25]{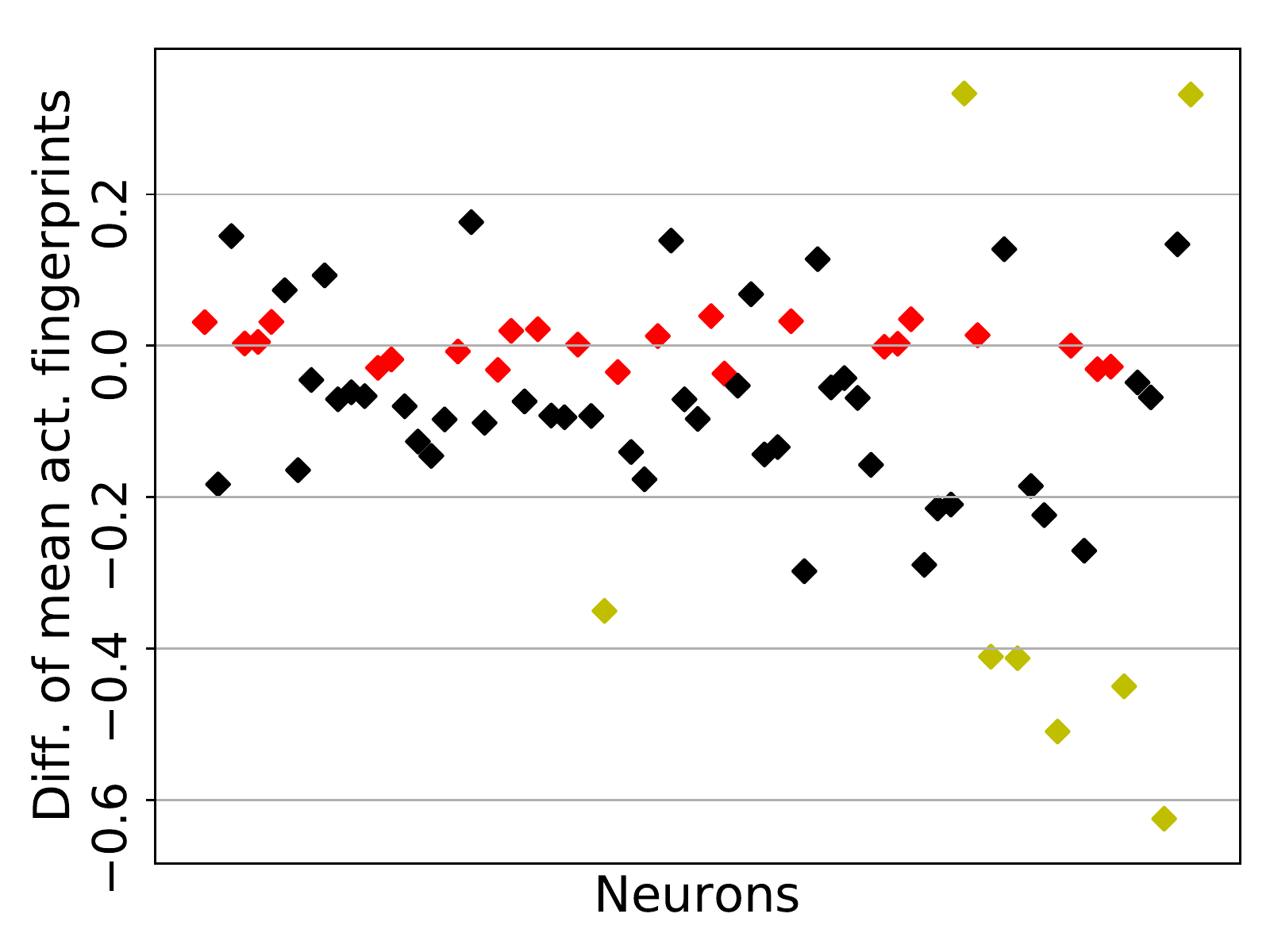}
\vspace{-1.5em}
\caption{Differences in neuron activation values between normal and
  adversarial inputs.}
\vspace{-1.5em}
\label{plot:idea}
\end{figure}

\begin{figure*}[t]
\includegraphics[scale=0.45, clip,trim=0.5cm 0.5cm 0.5cm 0.5cm]{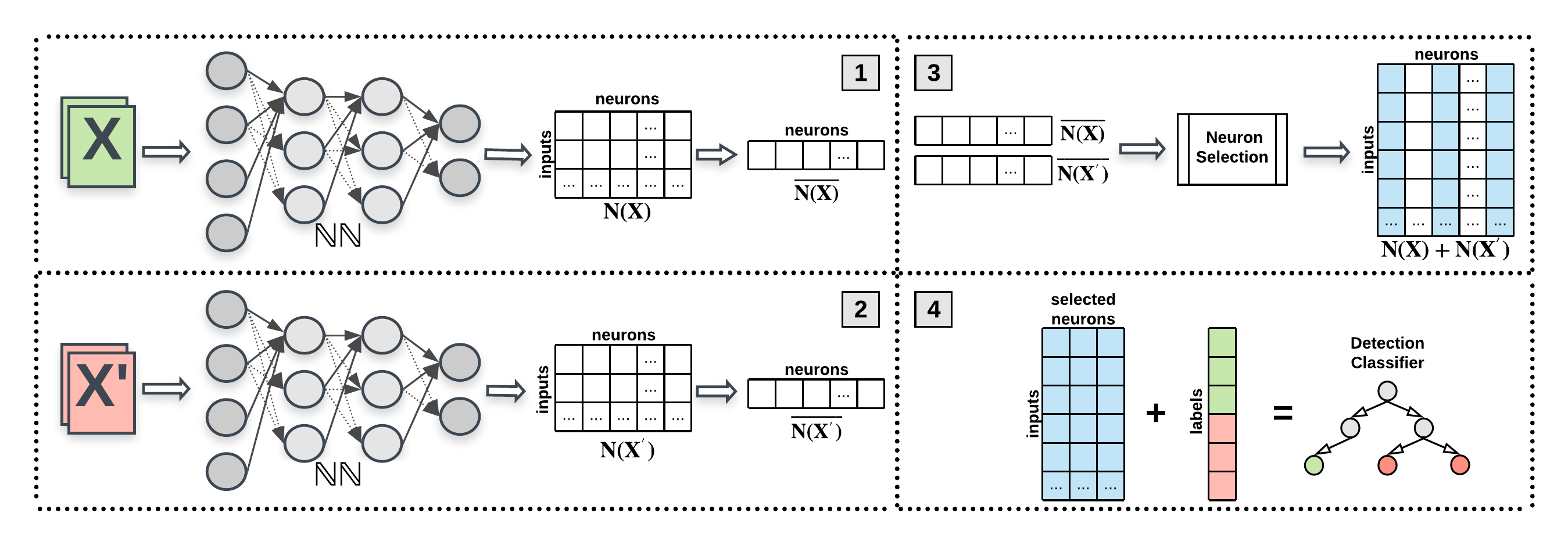}
\vspace{-1em}
\caption{Overview of \tool.}
\vspace{-1em}
\label{plot:overview}
\end{figure*}


Correctly classified (or normal) and adversarial inputs might look
almost identical to the human eye. However, the \emph{activation
  fingerprints}, that is, the neuron activation values, that
these two kinds of inputs leave on a neural network are certainly
different. This is because adversarial inputs cause the neural network
to make a different prediction, and this difference is manifested
through a change in the activation values of the neurons in the output
network layer. For the activation values of these neurons to be
different, there must also be changes in the activation values of
neurons in previous layers.

As an example, we train a neural network on the MNIST~\cite{MNIST}
dataset for identifying handwritten-digit images. We then collect the
activation values of four arbitrarily selected neurons for both normal
and adversarial images, which are generated by the PGD attack. On the
left of \figref{plot:idea}, we draw a box-plot of their activation
values distinguishing between normal and adversarial images. As shown
in the figure, there is a clear difference in the activation values of
these neurons for the two kinds of images. In this paper, we present a
Randomized Adversarial-Image Detection approach, which we call \tool,
leveraging exactly this difference.

\figref{plot:overview} gives an overview of \tool. In step 1, \tool provides a set of
normal inputs to the neural network under threat, and for each input, records the
activation values of every neuron. It also calculates the mean activation value of every
neuron for all normal inputs. In step 2, \tool repeats this process for adversarial inputs
that are generated by perturbing the normal inputs of the first step. In step 3, \tool
selects a number of neurons to monitor based on the mean difference in activation
values. In step 4, \tool trains a detection classifier based on the recorded activation
values of the selected neurons.

To decide whether a new input is adversarial, \tool monitors the activation values of the
selected neurons during prediction to obtain the fingerprint that is fed to the trained
detection classifier.

\subsection{Definitions}

Let $\mathbf{N} = \{n_1, n_2, \dots\}$ be the set of all neurons
(excluding input and output layers) in a neural network $\mathbb{NN}$
and $X=\{x_1, x_2, \dots\}$ an arbitrary set of inputs. We denote the
activation value of a neuron $n_i$ for an input $x_j$ as $n_i(x_j)$.

We call $n_i(X)$ the \emph{activation-value block} of $n_i$ with
respect to inputs $X$, which essentially denotes a vector of
activation values $n_i(x_j)$ for each input $x_j \in X$ (neuron $n_i$
is fixed).

$N$ stands for an ordered subset of all neurons $\mathbf{N}$ in the
neural network. $N(x_j)$ denotes the \emph{activation fingerprint}
(AF) of input $x_j$ on the neurons in subset $N$. The activation
fingerprint is defined as a vector of activation values $n_i(x_j)$ for
each neuron $n_i \in N$ (input $x_j$ is fixed).

$N(X)$ yields a two-dimensional matrix, where each row corresponds to
the activation fingerprint of an input $x_j \in X$ on neurons $N$ and
each column to the activation-value block of a neuron $n_i \in N$ with
respect to inputs $X$.

We define a \emph{mean activation fingerprint}, denoted
$\overline{N(X)}$, as a vector that replaces each
activation-value block in $N(X)$ by its mean activation value. Then,
the \emph{difference of mean activation fingerprints} for two sets of
inputs $X$ and $Y$ is defined as |$\overline{N(X)} -
\overline{N(Y)}$|, where the subtraction is performed element-wise.

\subsection{\tool}
\label{subsect:RAID}

\tool, is a simple, yet very effective, technique for detecting
adversarial images. It is based on the key insight of leveraging
differences in the activation fingerprints that normal and adversarial
inputs leave on the $\mathbb{NN}$ under threat.

To make use of this insight, \tool starts by computing $\mathbf{N}(X)$
and $\mathbf{N}(X')$ for all neurons in $\mathbb{NN}$, where $X$ is
a set of normal inputs, and $X'$ the set of adversarial inputs
generated by perturbing the inputs in $X$.
In our experience, however, not all neurons behave differently for
normal and adversarial inputs. For instance, it could be the case
that, for a particular neuron $n_i$, the activation-value blocks
$n_i(X)$ and $n_i(X')$ in $\mathbf{N}(X)$ and $\mathbf{N}(X')$,
respectively, contain similar activation values.

On the right of \figref{plot:idea}, we plot the difference
$\overline{\mathbf{N}(M)} - \overline{\mathbf{N}(M')}$ for all
neurons in a simple neural network
trained on the MNIST dataset.  Here,
$M$ denotes the set of all images in MNIST and $M'$ the adversarial
images generated by DF.
Each point in the figure corresponds to $\overline{n(M)} -
\overline{n(M')}$, that is, the difference of mean activation-value
blocks. Neurons with $|\overline{n(M)} - \overline{n(M')}|$ $<
0.03$ appear in red, neurons with $|\overline{n(M)} -
\overline{n(M')}| > 0.35$ in yellow, and the remaining neurons in
black. We call red neurons \emph{inessential} as their behavior does
not change significantly for images in $M$ and $M'$. Consequently, such neurons are
not useful for detecting adversarial examples. On the other hand,
yellow neurons have a very clear difference in behavior and should be
leveraged.


Our technique, therefore, filters out inessential neurons. More
specifically, \tool computes and sorts the vector
$|\overline{\mathbf{N}(X)} - \overline{\mathbf{N}(X')}|$. Then,
based on a given percentage, which is a hyper-parameter of our
technique called \emph{filtering threshold}, \tool drops the
percentage of this sorted vector with the lowest activation values.

\tool uses the activation values of the remaining \emph{essential}
neurons to train an adversarial-input detection classifier. In
particular,
our technique randomly selects a number
of essential neurons to monitor, which is also a hyper-parameter of
\tool. It then labels each AF of input $x \in X$ on the monitored
neurons as normal, and each AF of $x' \in X'$ as adversarial. These
AFs together with their labels are used to train our detection
classifier. In \secref{sect:experiments}, we show how the number of
monitored neurons impacts the effectiveness of our technique. We also
discuss why we \emph{randomly} select the monitored neurons, instead
of deterministically picking  them.

The detection classifier takes the AF that a new input of the neural
network leaves on the monitored neurons and decides whether the input
is adversarial. Due to the relatively small number of monitored
neurons, the input space of the detection classifier is not high
dimensional, that is, the AFs do not have too many features. As a
result, this allows us to choose a simple type of classifier without
compromising the effectiveness of our approach.
Instead of training a large detection neural network as in existing classifier-based work (e.g., \cite{MetzenGenewein2017}), our experiments demonstrate
the competitive effectiveness of much simpler classifiers, such as
decision trees, random forests, etc.

\subsection{P-\tool: Pooled \tool}
\label{subsect:P-RAID}

Existing classifier-based detection
techniques~\cite{MetzenGenewein2017,GongWang2017,GrosseManoharan2017}
have been very successful in detecting adversarial examples. Despite
their success, they have been criticized for being vulnerable to
adaptive white-box attacks~\cite{CarliniWagner2017-Bypassing}. In
particular, the criticism is that if a white-box attack can trick a
neural network into making a wrong prediction, then it should also be
able to bypass the detection classifier by adapting its adversarial
attacks.
We, therefore, extend our approach to be robust against adaptive
white-box attacks. We refer to our extended technique as P-\tool, for
Pooled-\tool.

The difference between \tool and P-\tool consists in training a
\emph{pool} of detection classifiers, instead of a single one. Each
classifier in the pool is trained with the AFs left on an equal number
of \emph{randomly} selected essential neurons. These neurons are
selected uniformly from the entire $\mathbb{NN}$ (after having
filtered out the inessential neurons). This results in distinct
classifiers, which however are trained for the same goal. We pick an
equal number of neurons for each classifier so that they are all
similarly effective in detecting adversarial examples---recall that
the number of monitored neurons impacts the effectiveness of the
classifiers.

Now, for each new input of the neural network, P-\tool selects
\emph{uniformly at random} a detection classifier from the pool. The
selected classifier determines if the input is
adversarial. As we show in \secref{sect:experiments}, extending our
technique in this way does not impact its effectiveness. On the
contrary, we argue that it improves its robustness against adaptive
white-box attacks. Given that P-\tool picks the detection
classifier \emph{nondeterministically} for each input, an adaptive attacker would have to tailor
its adversarial attacks to \emph{all classifiers in the pool}. This is
infeasible for three main reasons. First, the size of the pool can be
arbitrarily large. Second, an attack that is optimized against one
classifier might be impossible to also optimize against another since
they are trained on different sets of neurons. Third, the pool of
classifiers could contain different classifier types (e.g., k-nearest
neighbors, random forests, etc.), which makes it even more difficult
to tailor an attack to all classifiers.

\section{Experimental Evaluation}
\label{sect:experiments}

In this section, we address the following research questions using
established evaluation guidelines for adversarial
robustness~\cite{CarliniWagner2017-Bypassing,CarliniAthalye2019}:
\begin{description}
\item[RQ1:] How effective is our approach in detecting adversarial
  images?
\item[RQ2:] Does our approach generalize within and across attack
  norms?
\item[RQ3:] How does our approach compare with state-of-the-art
  detection techniques?
\item[RQ4:] How does the selection of monitored neurons impact the
  effectiveness of our approach?
\item[RQ5:] How do multiple detection classifiers impact the effectiveness of
  our approach?
\item[RQ6:] How effective are different detection classifier types?
\end{description}

\subsection{Implementation}
\label{subsect:implementation}

We implemented (P-)\tool in Python using the popular machine-learning
framework Keras~\cite{Keras} (v2.3.1) with the
Tensorflow~\cite{Tensorflow} (v1.15.0) back-end for analyzing neural
networks. We also employ the scikit-learn
library~\cite{PedregosaVaroquaux2011} for training detection
classifiers. Last, we use IBM's Adversarial Robustness Toolbox
(ART)~\cite{NicolaeSinn2018} for generating adversarial examples. Our
implementation is open source.

\subsection{Setup}
\label{subsect:setup}


We set up our experiments as follows.

\paragraph{\textbf{Datasets and network models.}}
We evaluate our technique on neural-network models trained on two
popular datasets, namely MNIST~\cite{MNIST} and
CIFAR-10~\cite{Krizhevsky2008}.
MNIST is a dataset for recognizing handwritten-digit images, whereas
the CIFAR-10 dataset focuses on recognizing objects and classifying
them into ten categories. It is common to evaluate the effectiveness
of adversarial-image detection techniques on these two datasets (e.g.,
\cite{KimFeldt2019,WangDong2019}).
For MNIST, we trained a 5-layer convolutional neural network
(ConvNet), with 320 neurons and 99.31\% accuracy (when using
60,000 images for training and 10,000 for testing). For CIFAR-10, we
trained a 12-layer ConvNet, with 2,208 neurons and 82.27\%
accuracy (when using 50,000 images for training and 10,000 for
testing). These specific network models have also been used to
evaluate a recent related technique~\cite{KimFeldt2019}.

\paragraph{\textbf{Adversarial images.}}
We generate adversarial images using six well-known attacks (see
\secref{subsect:attacks}). The hyper-parameter configuration of each attack is given below:
\begin{itemize}
\item \textbf{PGD}: This is an iterative $\text{L}_\infty$-norm attack. We set the maximum number
  of iterations to 100 and the maximum distortion to 0.3 in terms of the
  $\text{L}_\infty$ norm.
\item \textbf{FGSM}: This is a one step $\text{L}_\infty$-norm attack. We set the maximum
  distortion to 0.3 in terms of the $\text{L}_\infty$ norm.
\item \textbf{BIM}: This attack is the iterative version of the FGSM attack. The maximum
  number of iterations is set to 100 and the maximum distortion to 0.3 in terms of
  the $\text{L}_\infty$ norm.
\item \textbf{DF}: This is an iterative $\text{L}_2$-norm attack. We set the maximum number
  of iterations to 100.
\item \textbf{CW}: Although there is an $\text{L}_\infty$-norm version of this attack, we employ
  its stronger $\text{L}_2$-norm version~\cite{CarliniWagner2017-Robustness}. We use the default
  settings of the ART library~\cite{NicolaeSinn2018}.
\item \textbf{JSMA}: This is an $\text{L}_0$-norm attack. We set parameter $\mathit{gamma}$ bounding
  the fraction of perturbed features to 1.0.
\end{itemize}

We use each of these attacks to generate adversarial images in the
following way. For each dataset, we split the test set, that is, the
10,000 images, into its halves. From each set of 5,000 original
images, we remove any misclassified images and label the rest as
normal. Next, we use each of the above attacks to generate 5,000
adversarial images for each set of normal images. We discard any
generated images that are correctly classified and label the rest as
adversarial. We then use one set of normal images together with its
corresponding adversarial images to train a random-forest classifier
based on the activation values of the monitored neurons. The remaining
normal and adversarial images are used to test the ability of the
classifier to detect adversarial images.

\paragraph{\textbf{Existing detection techniques.}}
In our experiments, we compare our approach with three
state-of-the-art adversarial-image detection techniques from the
literature~\cite{KimFeldt2019,WangDong2019}, two of which are
classifier based and one is not.

Kim et
al.~\cite{KimFeldt2019} present two techniques based on the \emph{surprise measure} of a
given input to a neural network. They show that this surprise measure can be used to
detect adversarial inputs, under the assumption that such inputs are more
surprising than normal ones. The first technique presented in their paper, namely
Likelihood-based Surprise Adequacy (LSA), is an extension of previous
work~\cite{FeinmanCurtin2017}. In LSA, they estimate the probability density function of
each neuron's activation values, which is then used to measure how surprising an
incoming input is. The second technique, namely Distance-based Surprise Adequacy (DSA), is a
novel method based on the activation values of a set of neurons (i.e., a layer of
neurons). In DSA, the surprise of an input $x$ is measured by the distances between the activation-value
vectors of the selected neurons from  the closest input $y$ in the same class and between $y$ and the
closest input $z$ in a different class. Both LSA and DSA train a classifier based on surprise
values to distinguish adversarial and normal examples.

The third technique, mMutant~\cite{WangDong2019}, is based on the observation that, compared to a normal
example, it is easier to change the class of an adversarial example via small mutations to the
neural network. They first define the \emph{label change rate} (LCR) of an input on
slightly mutated versions of the original neural network. The LCR is found by dividing the
number of predictions on mutated models deviating from the original prediction over the
number of mutated models. The fundamental assumption behind mMutant is that the LCR
of adversarial inputs is higher than the LCR of normal ones. We investigated the
validity of this assumption, and our experiments suggest that it does not necessarily hold
in general. Their paper provides a detection algorithm based on this assumption that tries
to identify an LCR threshold for distinguishing between normal and adversarial inputs.
However, this algorithm yields worse results~\cite{WangDong2019} when compared to the AUC
score achieved by the LCRs directly. For this reason, we do not consider it here.

For selecting hyper-parameters for the above techniques, we pick the
configuration that achieves the best results. For LSA and DSA, we pick
the best layers as proposed in the original paper since we use the
same neural networks in our evaluation. For the attacks that are not
included in the original paper, we still try to pick the layer that is
the most effective. Note that, by re-optimizing hyper-parameters for
each experiment, we give these two techniques a significant advantage
over others (incl. \tool). The authors of mMutant describe four
mutation operators for neural networks. We limit our comparison to
Neuron Activation Inverse (NAI), which the authors show to be most
effective.

\paragraph{\textbf{Machine specifications.}}
We conducted all our experiments on a 32-core \mbox{Intel}
  \textregistered~Xeon \textregistered~Gold 6134M CPU~@~3.20GH
  machine with 768GB of memory, running Debian v10.

\subsection{Metrics}
\label{subsect:metrics}

We evaluate the effectiveness of our approach using the following
metrics.

\paragraph{\textbf{Detection accuracy.}}
The accuracy of a detection technique is a standard metric. For a set
containing both normal and adversarial images, accuracy is the
percentage of all images that are correctly detected as normal or
adversarial. Higher accuracy suggests a better detection technique.

\paragraph{\textbf{True-positive rate.}}
We refer to the adversarial images that are correctly identified by
our technique as \emph{true positives} (TP), and to those that are
incorrectly identified as \emph{false negatives} (FN). The
true-positive rate (TPR) is defined as $\frac{\text{TP}}{\text{TP} +
  \text{FN}}$ and denotes the ratio of correctly detected adversarial
images over all adversarial images. A higher true-positive rate is
better.

\paragraph{\textbf{False-positive rate.}}
We refer to the normal images that are correctly identified by our
technique as \emph{true negatives} (TN), and to those that are
incorrectly identified as \emph{false positives} (FP). The
false-positive rate (FPR) is defined as $\frac{\text{FP}}{\text{FP} +
  \text{TN}}$ and denotes the ratio of normal images that are detected
as adversarial over all normal images. A lower false-positive rate is
better.

\paragraph{\textbf{Area under curve of receiver operator characteristic.}}
The receiver operator characteristic (ROC) curve plots the TPR against
the FPR for all classification thresholds, which define the
classification boundary between classes. Assuming that a detection
classifier is expected to rank adversarial images higher than normal
ones, the area under the ROC curve (AUC) denotes the probability that
a randomly chosen adversarial image is ranked higher than a randomly
chosen normal image. An AUC score that is closer to 1 is better.

\subsection{Results}
\label{subsect:results}

We now present our experimental results for each of the above research
questions.

Unless stated otherwise, we configured our implementation to use
exactly one classifier (not a pool), which is a random forest with 32
decision trees (or estimators). After filtering out 50\% of the
neurons in the networks for MNIST and CIFAR-10, the random forest is
trained with the activation values of 64 randomly selected neurons.

We repeated all of our experiments 8 times, each time using a
different random seed for the classifier and a different randomly
selected set of neurons. Unless we explicitly state otherwise, the
rest of this section reports mean results and their standard
deviation.

\paragraph{\textbf{RQ1: Adversarial-image detection.}}
To measure the effectiveness of our approach in detecting adversarial
images, we evaluate it against several well-known
attacks. \tabsref{tab:rq1_mnist} and \ref{tab:rq1_cifar} show the
results for the MNIST and CIFAR-10 datasets, respectively.

\begin{table*}[t]
\caption{Effectiveness of \tool in detecting adversarial images
  generated by six attacks for MNIST.}
\vspace{-1em}
\label{tab:rq1_mnist}
\scalebox{0.95}{
\begin{tabular}{@{}lS[table-format=1.2]@{${}\pm{}$}S[table-format=1.2]S[table-format=1.2]@{${}\pm{}$}S[table-format=1.2]S[table-format=1.2]@{${}\pm{}$}S[table-format=1.2]S[table-format=1.2]@{${}\pm{}$}S[table-format=1.2]S[table-format=5.2]@{${}\pm{}$}S[table-format=2.2]S[table-format=3.2]@{${}\pm{}$}S[table-format=2.2]S[table-format=4.2]@{${}\pm{}$}S[table-format=2.2]S[table-format=3.2]@{${}\pm{}$}S[table-format=2.2]@{}}
\toprule
& \multicolumn{2}{c}{\textbf{Accuracy}} & \multicolumn{2}{c}{\textbf{TPR}} & \multicolumn{2}{c}{\textbf{FPR}} & \multicolumn{2}{c}{\textbf{AUC}} & \multicolumn{2}{c}{\textbf{TP}} & \multicolumn{2}{c}{\textbf{FP}} & \multicolumn{2}{c}{\textbf{TN}} & \multicolumn{2}{c}{\textbf{FN}} \\ \midrule
$\mathbf{L_*}$ &  0.96 & 0.00   & 0.99 & 0.00 &  0.16 & 0.02  &  0.99 & 0.00  &  23482.62 & 39.76  &  813.50 & 71.44 &  4089.50 & 71.44  &  220.38 & 39.76  \\
$\mathbf{L_{\infty}}$ & 1.00 & 0.00 & 1.00 & 0.00 & 0.00 & 0.00 & 1.00 & 0.00 & 14526.75 & 2.11 & 9.38 & 8.75 & 4893.62 & 8.75  & 2.25 & 2.11   \\
$\mathbf{L_2}$ &   0.94 & 0.00  & 0.96 & 0.01 & 0.09 & 0.01  & 0.98 & 0.00  &  8784.75 & 44.74 & 423.50 & 21.51  & 4479.50  & 21.51  &  369.25 & 44.74  \\
\textbf{PGD}  & 1.00 & 0.00 & 1.00 & 0.00 & 0.00 & 0.00 & 1.00 & 0.00 & 4969.62 & 0.86 & 5.75 & 6.56  &  4897.25  & 6.56 & 1.38 & 0.86  \\
\textbf{FGSM} & 1.00 & 0.00 & 1.00 & 0.00 & 0.00 & 0.00 & 1.00 & 0.00 &  4617.00 & 3.12 &  1.00  & 1.12 & 4902.00  & 1.12 & 2.00  & 3.12  \\
\textbf{BIM}  & 1.00 & 0.00 & 1.00 & 0.00 & 0.00 & 0.00 & 1.00 & 0.00  & 4936.38 & 3.00 &  2.88 & 4.73 & 4900.12 & 4.73 &  2.62 & 3.00  \\
\textbf{DF}   & 0.94 & 0.00 & 0.88 & 0.01 & 0.01  & 0.00 & 0.99  & 0.00 & 3709.25 & 34.23  &  68.25 & 25.55 & 4834.75 & 25.55 & 511.75 & 34.23  \\
\textbf{CW}   &  0.94  & 0.01 & 0.93 & 0.03  & 0.05 & 0.01 & 0.98 & 0.01 & 4633.50  & 59.42 & 284.62 & 13.90 & 4618.37  & 13.90 & 299.50 & 59.42 \\
\textbf{JSMA} & 0.95 & 0.00 & 0.95  & 0.01 & 0.06  & 0.00 &  0.99 & 0.00  &  4690.25 & 9.87 &  285.38 & 22.53 & 4617.62 & 22.53  & 220.75 & 9.87 \\ \bottomrule
\end{tabular}}
\vspace{-0.5em}
\end{table*}

\begin{table*}[t]
\caption{Effectiveness of \tool in detecting adversarial images
  generated by six attacks for CIFAR-10.}
\vspace{-1em}
\scalebox{0.95}{
\label{tab:rq1_cifar}
\begin{tabular}{@{}lS[table-format=1.2]@{${}\pm{}$}S[table-format=1.2]S[table-format=1.2]@{${}\pm{}$}S[table-format=1.2]S[table-format=1.2]@{${}\pm{}$}S[table-format=1.2]S[table-format=1.2]@{${}\pm{}$}S[table-format=1.2]S[table-format=5.2]@{${}\pm{}$}S[table-format=2.2]S[table-format=4.2]@{${}\pm{}$}S[table-format=2.2]S[table-format=4.2]@{${}\pm{}$}S[table-format=2.2]S[table-format=3.2]@{${}\pm{}$}S[table-format=2.2]@{}}
\toprule
& \multicolumn{2}{c}{\textbf{Accuracy}} & \multicolumn{2}{c}{\textbf{TPR}} & \multicolumn{2}{c}{\textbf{FPR}} & \multicolumn{2}{c}{\textbf{AUC}} & \multicolumn{2}{c}{\textbf{TP}} & \multicolumn{2}{c}{\textbf{FP}} & \multicolumn{2}{c}{\textbf{TN}} & \multicolumn{2}{c}{\textbf{FN}} \\ \midrule
$\mathbf{L_*}$ & 0.95 & 0.00 & 0.99 & 0.00 & 0.34 & 0.02 & 0.96 & 0.00 & 26346.56 & 8.67 &  1230.78 &  34.95  & 2360.22 & 34.95 &  296.44  & 8.67  \\
$\mathbf{L_{\infty}}$ & 1.00 & 0.00  & 1.00 & 0.00  & 0.00 & 0.00  & 1.00 & 0.00  & 14526.75 & 2.11 & 9.38 & 8.75 & 4893.62 & 8.75 & 2.25 & 2.11 \\
$\mathbf{L_2}$ & 0.88 & 0.00 & 0.96 & 0.00 & 0.30 & 0.01 & 0.90 & 0.00 &  7744.25 & 9.27 &  1089.62 & 30.60 &  2504.38 & 30.60 &  358.75 & 9.27 \\
\textbf{PGD}  & 1.00 & 0.00 & 1.00 & 0.00  & 0.00 & 0.00 & 1.00 & 0.00 & 4781.25 & 0.83 &  0.50 & 0.71  & 3587.50 & 0.71 &  0.75 & 0.83\\
\textbf{FGSM} & 1.00 & 0.00 & 1.00 & 0.00  & 0.00 & 0.00 & 1.00 & 0.00 & 4289.00 & 0.00 &  0.00 & 0.00  & 3588.00 & 0.00 &  0.00 & 0.00  \\
\textbf{BIM}  &  1.00  &  0.00   &  1.00   &  0.00   &  0.00  &  0.00  &  1.00  & 0.00   &  3588.00  &  0.00  & 0.00 & 0.00 & 0.00  & 0.00 & 4631.00 & 0.00 \\
\textbf{DF}   &  0.83 & 0.00 & 0.87 & 0.00 & 0.20 & 0.00 & 0.91 & 0.00 & 3470.00 & 17.36 & 719.88 & 19.76 & 2874.12 & 19.76 &  535.00 & 17.36  \\
\textbf{CW}   & 0.85 & 0.00 & 0.93  & 0.01 &  0.24 & 0.01 & 0.91 & 0.01 & 3816.25 &  14.62 &  842.50 & 32.76 &  2745.50 & 32.76 &  281.75 & 14.62   \\
\textbf{JSMA} & 0.90 & 0.00 &  0.93 & 0.01 & 0.13 & 0.01 & 0.96 & 0.00 & 4470.75 & 31.95  & 481.38 & 25.27  & 3106.62 & 25.27  &  364.25 & 31.95 \\ \bottomrule
\end{tabular}}
\vspace{-1em}
\end{table*}

\begin{table}[b]
\vspace{-1em}
\caption{AUC scores achieved by \tool when trained with all and tested
  on a subset of attacks.}
\vspace{-1em}
\label{tab:rq1alltox}
\scalebox{0.95}{
\begin{tabular}{@{}lccccccccc@{}}
\toprule
& \multicolumn{1}{c}{$\mathbf{L_{\infty}}$} & \multicolumn{1}{c}{$\mathbf{L_2}$} & \multicolumn{1}{c}{\textbf{PGD}} & \multicolumn{1}{c}{\textbf{FGSM}} & \multicolumn{1}{c}{\textbf{BIM}} & \multicolumn{1}{c}{\textbf{DF}} & \multicolumn{1}{c}{\textbf{CW}} & \multicolumn{1}{c}{\textbf{JSMA}} \\ \midrule
\textsc{\textbf{MNIST}} & 1.00& 0.97 & 1.00 & 1.00 & 1.00 & 0.99 & 0.97 & 0.98 \\
\textsc{\textbf{CIFAR-10}} & 1.00 & 0.89 & 1.00 & 1.00 & 1.00 & 0.90 & 0.89 & 0.94  \\ \bottomrule
\end{tabular}}
\end{table}

The first column of the tables indicates the used attack. For
instance, to obtain the results of the PGD row, we trained and tested
our classifier using two different sets, each containing $\sim$5,000
normal images as well as $\sim$5,000 adversarial ones, which were
generated by the PGD attack. Note that the number of images is
approximate due to filtering, which is described in detail in
\secref{subsect:setup}. For $\text{L}_2$, we trained and tested using
$\sim$5,000 normal images, $\sim$5,000 adversarial images generated by
DF, and $\sim$5,000 adversarial images generated by CW. For
$\text{L}_\infty$, we used PGD, FGSM, and BIM to generate adversarial
images, and for $\text{L}_*$, all six attacks. Note that an
$\text{L}_0$ row corresponds to the JSMA one as JSMA is the only
$\text{L}_0$ attack used in our experiments.  The remaining columns of
the tables show the detection accuracy, the true- and false-positive
rates, the area under the ROC curve, the number of true and false
positives as well as the number of true and false negatives.

As shown in \tabsref{tab:rq1_mnist} and \ref{tab:rq1_cifar}, \tool is
most effective in detecting $\text{L}_\infty$ attacks, with accuracy,
TPR, and AUC at 1.00 and FPR at 0.00, which constitute the theoretical
best. \tool is least effective for $\text{L}_2$ attacks; this is to be
expected since these are more
powerful~\cite{CarliniWagner2017-Bypassing}. Naturally, the
effectiveness for $\text{L}_*$ lies in-between.

\tabref{tab:rq1alltox} shows the AUC score achieved by \tool when
trained with normal and adversarial images generated by all attacks
($\text{L}_*$) and tested on normal and adversarial images generated
by the attacks shown in the columns of the table. For example, when
training \tool with all attacks, it achieves an AUC score of 1.0 when
tested only on FGSM attacks. Again, our approach is most effective in
detecting $\text{L}_\infty$ attacks and slightly less effective for
others.

Since \tool detects an adversarial image based on the prediction of a
classifier, its running time for a single input is in the order of milliseconds.
Training the classifier(s) used by (P-)\tool takes longer, but this
process is performed once and offline. Training time depends on three
factors: (1)~number of training inputs, (2)~number of features in each
input, and (3)~internal complexity of the classifier(s) (e.g., number of
estimators in a random forest). For \tool's best configuration
described at the beginning of this subsection, training a random
forest with 10,000 images takes less than 30 seconds.

\paragraph{\textbf{RQ2: Attack norms.}}
We refer to $\text{L}_0$, $\text{L}_2$, $\text{L}_\infty$, and
$\text{L}_*$ as attack norms. This research question focuses on
evaluating whether \tool's detection generalizes within and across
attack norms.

Within an attack norm, we train with (normal and) adversarial images
generated by an attack of a particular norm (e.g., PGD for
$\text{L}_\infty$, DF for $\text{L}_2$) and measure the AUC score
achieved by \tool when testing with (normal and) adversarial images
generated by a different attack of the same norm (e.g., FGSM for
$\text{L}_\infty$, CW for $\text{L}_2$). The results for
$\text{L}_\infty$ are shown in \tabref{tab:rq2_linf_x_to_y}, and for
$\text{L}_2$ in \tabref{tab:rq2_l2_x_to_y}. The first column of the
tables refers to the attack with which we train, whereas the first row
indicates the used dataset as well as the attack with which we test.

The results in \tabref{tab:rq2_linf_x_to_y} are impressive. \tool
achieves the best possible AUC score with 0.00 standard deviation for
all attacks within $\text{L}_\infty$. Therefore, our detection
technique generalizes perfectly within this norm. As shown in
\tabref{tab:rq2_l2_x_to_y} however, \tool does not generalize as well
within the $\text{L}_2$ norm. The AUC scores in this table are
slightly lower than those in \tabsref{tab:rq1_mnist} and
\ref{tab:rq1_cifar}, where we trained with $\text{L}_2$, DF, or CW and
tested with the same. This also holds when comparing with the scores
in \tabref{tab:rq1alltox}, where we trained with all attacks and
tested with $\text{L}_2$, DF, or CW. These results indicate that, for
the more powerful $\text{L}_2$ norm, our detection technique is more
effective when trained with the attack it is expected to detect.

\begin{table}[b]
\vspace{-1em}
\caption{AUC scores achieved by \tool when trained with a particular
  attack of the $\text{L}_\infty$ norm (shown in the first column) and
  tested with another of the same norm (first row).}
\vspace{-1em}
\label{tab:rq2_linf_x_to_y}
\scalebox{0.95}{
\begin{tabular}{c|ccc|ccc}
\toprule
\multicolumn{1}{l|}{} & \multicolumn{3}{c|}{\textbf{MNIST}}         & \multicolumn{3}{c}{\textbf{CIFAR-10}}         \\
                      & \textbf{PGD} & \textbf{FGSM} & \textbf{BIM} & \textbf{PGD} & \textbf{FGSM} & \textbf{BIM} \\ \midrule
\textbf{PGD}          & \cellcolor{black!50}  &  1.00 &  1.00 & \cellcolor{black!50}  & 1.00 & 1.00 \\
\textbf{FGSM}         & 1.00& \cellcolor{black!50}  &  1.00  & 1.00 & \cellcolor{black!50}   & 1.00 \\
\textbf{BIM}          & 1.00 &  1.00 & \cellcolor{black!50}  & 1.00 &   1.00 & \cellcolor{black!50} \\ \bottomrule
\end{tabular}}
\end{table}

Across attack norms, we train with (normal and) adversarial images
generated by all attacks of a particular norm (e.g., PGD, FGSM, and
BIM of $\text{L}_\infty$) and measure the AUC score achieved by \tool
when testing with (normal and) adversarial images generated by all
attacks of a different norm (e.g., DF and CW of $\text{L}_2$). The
results are shown in \tabref{tab:rq2_class_x_to_y}. The first column
of the table refers to the norm with which we train, whereas the first
row indicates the used dataset as well as the norm with which we test.

As previously observed (\tabref{tab:rq1alltox}), when training with
$\text{L}_*$, our technique is most effective in detecting
$\text{L}_\infty$ attacks and only slightly less effective for other
norms. When training with $\text{L}_2$, \tool's detection generalizes
quite well to all other attack norms ($\text{L}_*$, $\text{L}_\infty$,
and $\text{L}_0$). The same holds when training with $\text{L}_0$,
although this configuration is less effective for the CIFAR-10
dataset. These results suggest that training with stronger attacks
generalizes to detecting weaker ones. On the other hand, when training
with the weaker $\text{L}_\infty$ attacks, the AUC scores for the
detection of $\text{L}_2$ and $\text{L}_0$ attacks drop significantly.

\begin{table}[t]
\caption{AUC scores achieved by \tool when trained with a particular
  attack of the $\text{L}_2$ norm (shown in the first column) and
  tested with another of the same norm (first row).}
\vspace{-1em}
\label{tab:rq2_l2_x_to_y}
\scalebox{0.95}{
\begin{tabular}{c|cc|cc}
\toprule
\multicolumn{1}{l|}{} & \multicolumn{2}{c|}{\textbf{MNIST}} & \multicolumn{2}{c}{\textbf{CIFAR-10}} \\
                       & \textbf{DF}      & \textbf{CW}      & \textbf{DF}      & \textbf{CW}      \\  \midrule
\textbf{DF}            & \cellcolor{black!50}  &  0.88&  \cellcolor{black!50} &   0.86  \\
\textbf{CW}            & 0.88 & \cellcolor{black!50} & 0.86 & \cellcolor{black!50} \\  \bottomrule
\end{tabular}}
\vspace{-1.5em}
\end{table}

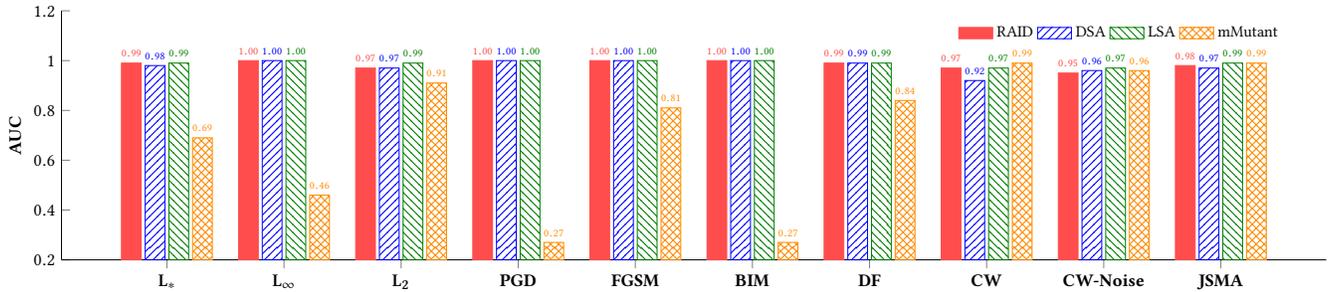
\begin{figure*}[t]
\centering
\scalebox{0.75}{
  \begin{tikzpicture}
    \begin{axis}[
      ybar,
      xticklabels={$\mathbf{L_*}$,$\mathbf{L_\infty}$,$\mathbf{L_2}$,\textbf{PGD},\textbf{FGSM},\textbf{BIM},\textbf{DF},\textbf{CW},\textbf{CW-Noise},\textbf{JSMA}},
      xtick=data,
      ylabel={\textbf{AUC}},
      legend cell align=left,
      axis y line*=none,
      axis x line*=bottom,
      width=24cm,
      height=6cm,
      ymin=0.2,
      ymax=1.2,
      ytick distance=0.2,
      bar width=0.35cm,
      area legend,
      every node near coord/.append style={font=\tiny},
      nodes near coords*={%
        \pgfmathprintnumber[fixed,fixed zerofill, precision=2]\pgfplotspointmeta},
      legend style={legend pos=north east,font=\small, draw=none, legend columns=-1}
      ]

      \pgfplotstableread{state-of-the-art-comparison-MNIST.dat} \comparison

      \addplot+[ybar, Red!70]
          table[x expr=\coordindex, y=RAID] from \comparison;

      \addplot+[ybar, Blue, pattern color=Blue, pattern=north east lines]
          table[x expr=\coordindex, y=DSA] from \comparison;

      \addplot+[ybar, Green, pattern color=Green, pattern=north west lines]
          table[x expr=\coordindex, y=LSA] from \comparison;

      \addplot+[ybar, DarkOrange, pattern color=DarkOrange, pattern=crosshatch]
          table[x expr=\coordindex, y=MUT] from \comparison;

      \legend{\tool, DSA, LSA, mMutant}
    \end{axis}
  \end{tikzpicture}
}
\vspace{-1em}
\caption{AUC scores of \tool, DSA, LSA, and mMutant on the MNIST
  dataset when tested on all attacks and norms.}
\vspace{-1em}
\label{fig:rq3MNIST}
\end{figure*}

\begin{figure*}[t]
\centering
\scalebox{0.75}{
  \begin{tikzpicture}
    \begin{axis}[
      ybar,
      xticklabels={$\mathbf{L_*}$,$\mathbf{L_\infty}$,$\mathbf{L_2}$,\textbf{PGD},\textbf{FGSM},\textbf{BIM},\textbf{DF},\textbf{CW},\textbf{CW-0.95},\textbf{CW-Noise},\textbf{JSMA}},
      xtick=data,
      ylabel={\textbf{AUC}},
      legend cell align=left,
      axis y line*=none,
      axis x line*=bottom,
      width=24cm,
      height=6.5cm,
      ymin=0,
      ymax=1.2,
      ytick distance=0.2,
      bar width=0.35cm,
      area legend,
      every node near coord/.append style={font=\tiny},
      nodes near coords*={%
        \pgfmathprintnumber[fixed,fixed zerofill, precision=2]\pgfplotspointmeta},
      legend style={legend pos=north east,font=\small, draw=none, legend columns=-1}
      ]

      \pgfplotstableread{state-of-the-art-comparison-CIFAR.dat} \comparison

      \addplot+[ybar, Red!70]
          table[x expr=\coordindex, y=RAID] from \comparison;

      \addplot+[ybar, Blue, pattern color=Blue, pattern=north east lines]
          table[x expr=\coordindex, y=DSA] from \comparison;

      \addplot+[ybar, Green, pattern color=Green, pattern=north west lines]
          table[x expr=\coordindex, y=LSA] from \comparison;

      \addplot+[ybar, DarkOrange, pattern color=DarkOrange, pattern=crosshatch]
          table[x expr=\coordindex, y=MUT] from \comparison;

      \legend{\tool, DSA, LSA, mMutant}
    \end{axis}
  \end{tikzpicture}
}
\vspace{-1em}
\caption{AUC scores of \tool, DSA, LSA, and mMutant on the CIFAR-10
  dataset when tested on all attacks and norms.}
\vspace{-1em}
\label{fig:rq3CIFAR}
\end{figure*}
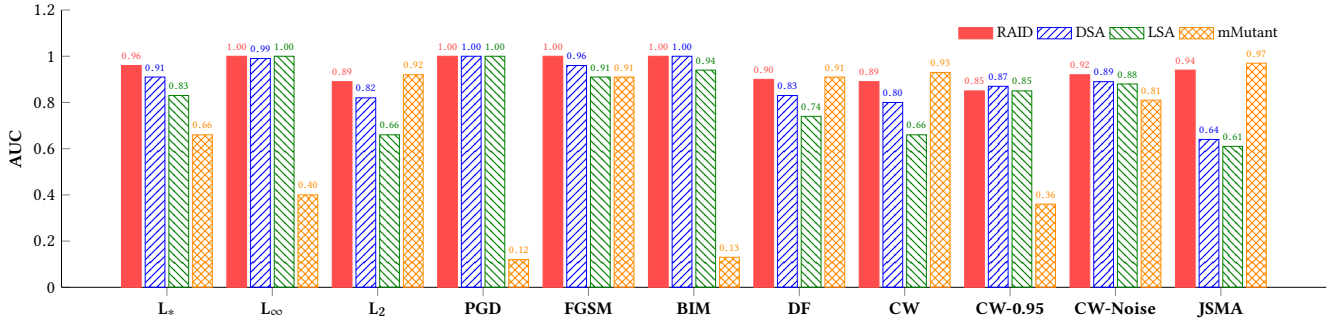

\paragraph{\textbf{RQ3: Comparison with state of the art.}}
We now compare \tool with the three state-of-the-art detection
techniques discussed in \secref{subsect:setup}, namely DSA, LSA, and
mMutant.

For this experiment, we measure the AUC score achieved by each
technique when tested with each attack and attack norm. The results
are shown in \figref{fig:rq3MNIST} for MNIST and \figref{fig:rq3CIFAR}
for CIFAR-10. For the techniques that require training, that is, for
\tool, DSA, and LSA, we train with $\text{L}_*$. We chose this
configuration because it is the most general, and thus, the most
natural. In practice, a detection technique is not a-priori aware of
the type or norm of future attacks; it is, therefore, safer to
anticipate all of them.

To obtain the AUC scores in the figures, we test each technique using
normal and adversarial images, which are generated by the attacks
shown on the x-axes of the bar charts. In addition to the aforementioned attacks,
we add two more for this experiment. First, we simulate an adaptive adversary
against mMutant by introducing CW-0.95, which is exactly the same as CW except that
the \emph{confidence} parameter is set to 0.95. In other words, CW-0.95 generates adversarial
images classified with a prediction confidence of at least 0.95 at the expense of
adding more distortion. Note that we only add this attack for CIFAR-10 since CW cannot
generate adversarial inputs with high confidence on
MNIST. Second, we introduce CW-Noise denoting that
our test set contains, in addition to the
$\sim$5,000 normal images, $\sim$5,000 adversarial images generated by
CW as well as $\sim$5,000 normal images with Gaussian noise (mean=0,
std=0.2) by following the guidelines presented in \cite{CarliniAthalye2019}.
We apply the filtering described in \secref{subsect:setup}
on the test sets of all techniques.

As shown in \figref{fig:rq3MNIST}, the effectiveness of \tool, DSA,
and LSA is comparable on the MNIST dataset. In the best case, \tool
outperforms DSA by 0.05 for CW, and LSA by 0.03 for CW-0.95. In the
worst case, LSA outperforms \tool by 0.02 for CW-Noise. The
effectiveness of mMutant varies significantly. For example, it
outperforms all other techniques for the powerful CW attacks, but is
substantially worse for the simple PGD attacks. This is because
mMutant is based on the assumption that adversarial examples are
sensitive to mutations, that is, they are typically classified with a
low prediction confidence. However, iterative attacks such as BIM and PGD generate adversarial images
with very high prediction confidence (e.g., 0.99 almost all the time),
which makes them more insensitive, and hence, harder to detect by
mMutant. In this sense, the more distortion is added to the input,
the lower the chance that mMutant detects it since higher distortion typically
increases prediction confidence.

\begin{table}[t]
\caption{AUC scores of \tool when trained with a particular norm
  (first column) and tested with another (first row).}
\vspace{-1em}
\label{tab:rq2_class_x_to_y}
\scalebox{0.95}{
\begin{tabular}{c|cccc|cccc}
\toprule
\multicolumn{1}{l|}{} & \multicolumn{4}{c|}{\textbf{MNIST}}       & \multicolumn{4}{c}{\textbf{CIFAR-10}}          \\
                      & $\mathbf{L_*}$ & $\mathbf{L_\infty}$ & $\mathbf{L_2}$ & $\mathbf{L_0}$ & $\mathbf{L_*}$ &$\mathbf{L_\infty}$ & $\mathbf{L_2}$ & $\mathbf{L_0}$ \\ \midrule
$\mathbf{L_*}$   & \cellcolor{black!50} & 1.00 & 0.97 & 0.98 &  \cellcolor{black!50} &  1.00 & 0.89 &  0.94 \\
$\mathbf{L_\infty}$ &  0.84 & \cellcolor{black!50} &  0.67 &  0.68 &  0.77 & \cellcolor{black!50} & 0.54  & 0.52  \\
$\mathbf{L_2}$  &  0.96 & 0.96 & \cellcolor{black!50} &  0.95 &  0.92 & 0.95 & \cellcolor{black!50} &  0.91 \\
$\mathbf{L_0}$ &  0.92 & 0.90 & 0.91 &\cellcolor{black!50} &  0.81 &  0.73 & 0.85 & \cellcolor{black!50}  \\ \bottomrule
\end{tabular}}
\vspace{-1em}
\end{table}

\figref{fig:rq3CIFAR} shows the same comparison for CIFAR-10. Here,
\tool consistently outperforms DSA and LSA, namely for $\text{L}_*$,
$\text{L}_\infty$, $\text{L}_2$, FGSM, BIM, DF, CW, CW-Noise, and
JSMA. Their effectiveness is equal for PGD, and DSA outperforms \tool
by 0.02 for CW-0.95. \tool outperforms mMutant for $\text{L}_*$,
$\text{L}_\infty$, PGD, FGSM, BIM, CW-0.95, and CW-Noise, in the best
case by 0.88 for PGD and in the worst case by 0.09 for FGSM. On the
other hand, mMutant looks slightly more effective than
\tool for $\text{L}_2$ (by 0.03), DF (by 0.01), CW (by 0.04), and JSMA
(by 0.03). However, as shown with CW-0.95, increasing prediction confidence for these
cases would reduce mMutant's effectiveness.

In addition to (often significantly) outperforming state-of-the-art
detection techniques, \tool is also the most consistent in terms of
effectiveness across different types of attacks and
norms. \tabref{tab:rq3ranges} shows the ranges of AUC scores achieved
by \tool, DSA, LSA, and mMutant for the attacks in
\figsref{fig:rq3MNIST} and~\ref{fig:rq3CIFAR}. As shown in the table,
the ranges are more stable for \tool than for any other technique.

In terms of performance, \tool is also very efficient. At runtime, \tool requires (1)
feeding the incoming input to the original neural network to collect its AF and (2) feeding
the AF to the detection classifier; in total, this takes less than a second as described
in RQ1. LSA, DSA, and mMutant demonstrate a similarly good runtime performance. mMutant requires
feeding the input to a number of mutated neural networks (default is 500).  Given that
each neural-network query takes less than a millisecond, mMutant can make a decision under 1 second. Note
that, although model mutation takes much more time (i.e., per mutation slightly more than 5
seconds for MNIST and slightly more than 15 seconds for CIFAR-10), it
is performed offline and incurs no runtime overhead.  Similarly to \tool, LSA and DSA involve
feeding the input to the original network and then feeding the extracted data to the
classifier; overall, these steps also take around 1 second.

\begin{figure}[b]
\vspace{-1em}
\begin{minipage}{4.2cm}
\centering
\scalebox{0.95}{
\begin{tikzpicture}
  \begin{axis}[
      xtick=data,
      xticklabels={1,4,16,64,256},
      width=4.5cm,
      height=4.5cm,
      xmin=1,
      xmax=5,
      ymin=0.4,
      ymax=1.1,
      ytick distance=0.2,
      xmajorgrids=true,
      ymajorgrids=true,
      legend style={legend pos=south east,font=\small, draw=none},
      xlabel=\textbf{Number of neurons},
      ylabel=\textbf{AUC score}]
    \addplot[mark=*,color=Red!70]
        plot coordinates {
        (1, 0.59)
        (2, 0.84)
        (3, 0.97)
        (4, 0.99)
        (5, 0.99)
    };
    \addlegendentry{$\mathbf{L_*}$}

    \addplot[color=Blue,mark=pentagon*]
        plot coordinates {
            (1, 0.74)
            (2, 0.94)
            (3, 1.0)
            (4, 1.0)
            (5, 1.0)
    };
    \addlegendentry{$\mathbf{L_\infty}$}

    \addplot[color=Green,mark=square*]
        plot coordinates {
            (1, 0.63)
            (2, 0.82)
            (3, 0.96)
            (4, 0.98)
            (5, 0.98)
    };
    \addlegendentry{$\mathbf{L_2}$}

    \addplot[color=DarkOrange,mark=triangle*]
        plot coordinates {
            (1, 0.59)
            (2, 0.83)
            (3, 0.97)
            (4, 0.99)
            (5, 0.99)
    };
    \addlegendentry{$\mathbf{L_0}$}
    \end{axis}
\end{tikzpicture}}
\end{minipage}%
\begin{minipage}{4.2cm}
\centering
\scalebox{0.95}{
\begin{tikzpicture}
  \begin{axis}[
      xtick=data,
      xticklabels={1,4,16,64,256},
      width=4.5cm,
      height=4.5cm,
      xmin=1,
      xmax=5,
      ymin=0.4,
      ymax=1.1,
      ytick distance=0.2,
      xmajorgrids=true,
      ymajorgrids=true,
      legend style={legend pos=south east,font=\small, draw=none},
      xlabel=\textbf{Number of neurons},
      ylabel=\textbf{AUC score}]

    \addplot[mark=*,color=Red!70]
        plot coordinates {
            (1, 0.58)
            (2, 0.78)
            (3, 0.93)
            (4, 0.96)
            (5, 0.96)
    };
    \addlegendentry{$\mathbf{L_*}$}

    \addplot[color=Blue,mark=pentagon*]
        plot coordinates {
            (1, 0.94)
            (2, 0.97)
            (3, 1.00)
            (4, 1.00)
            (5, 1.00)
    };
    \addlegendentry{$\mathbf{L_\infty}$}

    \addplot[color=Green,mark=square*]
        plot coordinates {
            (1, 0.53)
            (2, 0.66)
            (3, 0.85)
            (4, 0.90)
            (5, 0.91)
    };
    \addlegendentry{$\mathbf{L_2}$}

    \addplot[color=DarkOrange,mark=triangle*]
        plot coordinates {
            (1, 0.56)
            (2, 0.76)
            (3, 0.92)
            (4, 0.96)
            (5, 0.97)
    };
    \addlegendentry{$\mathbf{L_0}$}
    \end{axis}
\end{tikzpicture}}
\end{minipage}
\vspace{-1em}
\caption{AUC scores of \tool for all attack norms versus the
  number of neurons (left: MNIST, right: CIFAR-10).}
\label{fig:rq4Neurons}
\end{figure}
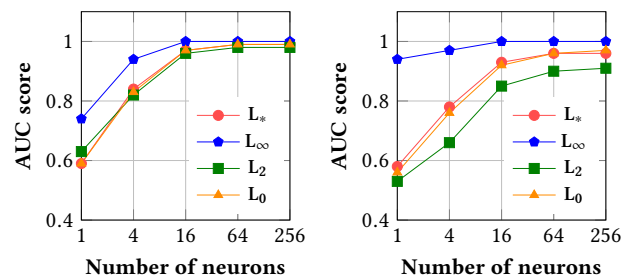

\paragraph{\textbf{RQ4: Selection of monitored neurons.}}
To determine how the selection of monitored neurons impacts the
effectiveness of our technique, we first measure the AUC scores
achieved by \tool for different numbers of monitored neurons, namely
1, 4, 16, 64, and 256 (or all essential neurons if their total number
is smaller than 256). For this experiment, we train and test \tool on
each attack norm. The results are presented in
\figref{fig:rq4Neurons}, where the plot on the left is for MNIST and
the plot on the right for CIFAR-10.

As shown in the figure, more neurons lead to better AUC scores,
although the differences are insignificant when comparing 64 and 256
neurons. This is expected because the larger the number of
monitored neurons, the higher are the chances of correctly detecting
more normal and adversarial examples. In particular, the activation
value of a neuron may fluctuate only for a specific set of normal and
adversarial images, whereas the activation value of another neuron may
fluctuate for a completely disjoint set. Therefore, when monitoring
both of these neurons, our technique is likely to be more
effective. In practice, with only a single neuron to monitor, \tool
may be lucky with its selection as for $\text{L}_\infty$ and CIFAR-10
in \figref{fig:rq4Neurons}, but this is typically not the case as is
also shown in the figure.

\begin{table}[b]
\vspace{-1em}
\caption{Range of AUC scores achieved by \tool, DSA, LSA, and mMutant
  when tested on the attacks of \figsref{fig:rq3MNIST}
  and~\ref{fig:rq3CIFAR}.}
\vspace{-1em}
\label{tab:rq3ranges}
\scalebox{0.95}{
\begin{tabular}{@{}lllll@{}}
\toprule
& \multicolumn{1}{c}{\textbf{\tool}} & \multicolumn{1}{c}{\textbf{DSA}} & \multicolumn{1}{c}{\textbf{LSA}} & \multicolumn{1}{c}{\textbf{mMutant}} \\ \midrule
\textsc{\textbf{MNIST}} & 0.95--1.00 & 0.92--1.00 & 0.93--1.00 & 0.27--0.99 \\
\textsc{\textbf{CIFAR-10}} & 0.85--1.00 & 0.64--1.00 & 0.61--1.00 & 0.12--0.97 \\ \bottomrule
\end{tabular}}
\end{table}

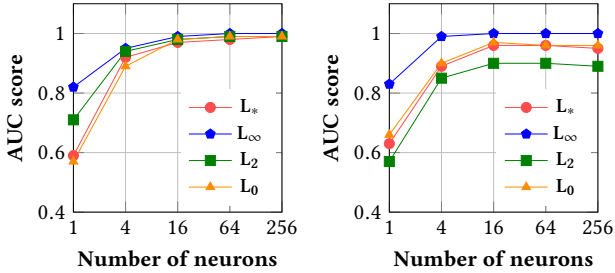
\begin{figure}[t]
\begin{minipage}{4.2cm}
\centering
\scalebox{0.95}{
\begin{tikzpicture}
  \begin{axis}[
      xtick=data,
      xticklabels={1,4,16,64,256},
      width=4.5cm,
      height=4.5cm,
      xmin=1,
      xmax=5,
      ymin=0.4,
      ymax=1.1,
      ytick distance=0.2,
      xmajorgrids=true,
      ymajorgrids=true,
      legend style={legend pos=south east,font=\small, draw=none},
      xlabel=\textbf{Number of neurons},
      ylabel=\textbf{AUC score}]
    \addplot[mark=*,color=Red!70]
        plot coordinates {
        (1, 0.59)
        (2, 0.92)
        (3, 0.97)
        (4, 0.98)
        (5, 0.99)
    };
    \addlegendentry{$\mathbf{L_*}$}

    \addplot[color=Blue,mark=pentagon*]
        plot coordinates {
            (1, 0.82)
            (2, 0.95)
            (3, 0.99)
            (4, 1.00)
            (5, 1.00)
    };
    \addlegendentry{$\mathbf{L_\infty}$}

    \addplot[color=Green,mark=square*]
        plot coordinates {
            (1, 0.71)
            (2, 0.94)
            (3, 0.98)
            (4, 0.99)
            (5, 0.99)
    };
    \addlegendentry{$\mathbf{L_2}$}

    \addplot[color=DarkOrange,mark=triangle*]
        plot coordinates {
            (1, 0.57)
            (2, 0.89)
            (3, 0.98)
            (4, 0.99)
            (5, 0.99)
    };
    \addlegendentry{$\mathbf{L_0}$}
    \end{axis}
\end{tikzpicture}}
\end{minipage}%
\begin{minipage}{4.2cm}
\centering
\scalebox{0.95}{
\begin{tikzpicture}
  \begin{axis}[
      xtick=data,
      xticklabels={1,4,16,64,256},
      width=4.5cm,
      height=4.5cm,
      xmin=1,
      xmax=5,
      ymin=0.4,
      ymax=1.1,
      ytick distance=0.2,
      xmajorgrids=true,
      ymajorgrids=true,
      legend style={legend pos=south east,font=\small, draw=none},
      xlabel=\textbf{Number of neurons},
      ylabel=\textbf{AUC score}]

    \addplot[mark=*,color=Red!70]
        plot coordinates {
            (1, 0.63)
            (2, 0.89)
            (3, 0.96)
            (4, 0.96)
            (5, 0.95)
    };
    \addlegendentry{$\mathbf{L_*}$}

    \addplot[color=Blue,mark=pentagon*]
        plot coordinates {
            (1, 0.83)
            (2, 0.99)
            (3, 1.00)
            (4, 1.00)
            (5, 1.00)
    };
    \addlegendentry{$\mathbf{L_\infty}$}

    \addplot[color=Green,mark=square*]
        plot coordinates {
            (1, 0.57)
            (2, 0.85)
            (3, 0.90)
            (4, 0.90)
            (5, 0.89)
    };
    \addlegendentry{$\mathbf{L_2}$}

    \addplot[color=DarkOrange,mark=triangle*]
        plot coordinates {
            (1, 0.66)
            (2, 0.90)
            (3, 0.97)
            (4, 0.96)
            (5, 0.96)
    };
    \addlegendentry{$\mathbf{L_0}$}
    \end{axis}
\end{tikzpicture}}
\end{minipage}
\vspace{-1em}
\caption{AUC scores of \tool for all norms when monitoring the best
  neurons (left: MNIST, right: CIFAR-10).}
\vspace{-1em}
\label{fig:rq4GoodNeurons}
\end{figure}

\begin{figure}[b]
\vspace{-1em}
\begin{minipage}{4.2cm}
\centering
\scalebox{0.95}{
\begin{tikzpicture}
  \begin{axis}[
      xtick=data,
      xticklabels={1,4,16,64,256},
      width=4.5cm,
      height=4.5cm,
      xmin=1,
      xmax=5,
      ymin=0.4,
      ymax=1.1,
      ytick distance=0.2,
      xmajorgrids=true,
      ymajorgrids=true,
      legend style={legend pos=south east,font=\small, draw=none},
      xlabel=\textbf{Number of neurons},
      ylabel=\textbf{AUC score}]
    \addplot[mark=*,color=Red!70]
        plot coordinates {
        (1, 0.50)
        (2, 0.51)
        (3, 0.93)
        (4, 0.99)
        (5, 1.00)
    };
    \addlegendentry{$\mathbf{L_*}$}

    \addplot[color=Blue,mark=pentagon*]
        plot coordinates {
            (1, 0.50)
            (2, 0.50)
            (3, 1.00)
            (4, 1.00)
            (5, 1.00)
    };
    \addlegendentry{$\mathbf{L_\infty}$}

    \addplot[color=Green,mark=square*]
        plot coordinates {
            (1, 0.50)
            (2, 0.53)
            (3, 0.76)
            (4, 0.95)
            (5, 0.99)
    };
    \addlegendentry{$\mathbf{L_2}$}

    \addplot[color=DarkOrange,mark=triangle*]
        plot coordinates {
            (1, 0.50)
            (2, 0.50)
            (3, 0.97)
            (4, 1.00)
            (5, 1.00)
    };
    \addlegendentry{$\mathbf{L_0}$}
    \end{axis}
\end{tikzpicture}}
\end{minipage}%
\begin{minipage}{4.2cm}
\centering
\scalebox{0.95}{
\begin{tikzpicture}
  \begin{axis}[
      xtick=data,
      xticklabels={1,4,16,64,256},
      width=4.5cm,
      height=4.5cm,
      xmin=1,
      xmax=5,
      ymin=0.4,
      ymax=1.1,
      ytick distance=0.2,
      xmajorgrids=true,
      ymajorgrids=true,
      legend style={legend pos=north east,font=\small, draw=none},
      xlabel=\textbf{Number of neurons},
      ylabel=\textbf{AUC score}]

    \addplot[mark=*,color=Red!70]
        plot coordinates {
            (1, 0.50)
            (2, 0.50)
            (3, 0.50)
            (4, 0.50)
            (5, 0.50)
    };
    \addlegendentry{$\mathbf{L_*}$}

    \addplot[color=Blue,mark=pentagon*]
        plot coordinates {
            (1, 0.50)
            (2, 0.50)
            (3, 0.50)
            (4, 0.50)
            (5, 0.50)
    };
    \addlegendentry{$\mathbf{L_\infty}$}

    \addplot[color=Green,mark=square*]
        plot coordinates {
            (1, 0.50)
            (2, 0.50)
            (3, 0.50)
            (4, 0.50)
            (5, 0.50)
    };
    \addlegendentry{$\mathbf{L_2}$}

    \addplot[color=DarkOrange,mark=triangle*]
        plot coordinates {
            (1, 0.50)
            (2, 0.50)
            (3, 0.50)
            (4, 0.50)
            (5, 0.50)
    };
    \addlegendentry{$\mathbf{L_0}$}
    \end{axis}
\end{tikzpicture}}
\end{minipage}
\vspace{-1em}
\caption{AUC scores of \tool for all norms when monitoring the worst
  neurons (left: MNIST, right: CIFAR-10).}
\label{fig:rq4BadNeurons}
\end{figure}
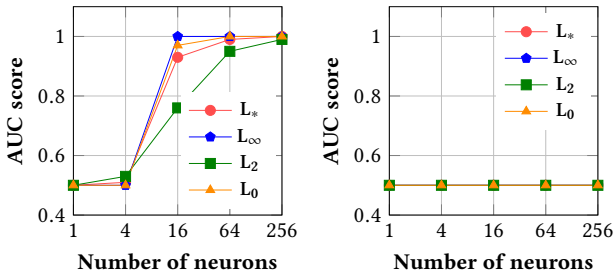

To further investigate the impact of monitored neurons on our
technique, we also performed the same experiment when selecting the
best neurons, that is, those with the largest difference of mean
activation fingerprints (see \secref{subsect:RAID}), and when selecting the
worst neurons. Note that the worst neurons are normally filtered out
by our technique as inessential. The results are shown in
\figsref{fig:rq4GoodNeurons} and~\ref{fig:rq4BadNeurons}. For this
experiment, unlike for the one in \figref{fig:rq4Neurons}, we do not
perform 8 runs of \tool since the sets of best and worst neurons are
chosen deterministically.

When comparing the AUC scores of \figsref{fig:rq4Neurons}
and~\ref{fig:rq4GoodNeurons}, we observe that selecting the best
neurons mostly benefits configurations with a small number of
neurons. For 64 neurons, which is the configuration used throughout
our experiments, \tool with the best neurons is better only for
$\text{L}_\infty$ and MNIST and only by 0.01, which justifies the
value of our filtering threshold (50\%). Interestingly, some AUC
scores achieved by \tool with the best neurons are slightly worse than
those with random neurons. This may happen when the activation values
of the best neurons fluctuate for overlapping sets of images, whereas
there are worse neurons that would help with the detection of other
images.

The AUC scores achieved by \tool when selecting the worst neurons
(\figref{fig:rq4BadNeurons}) are significantly worse than those in
\figsref{fig:rq4Neurons} and~\ref{fig:rq4GoodNeurons}, which justifies
the existence of our filtering threshold. For CIFAR-10, all AUC scores
are 0.5. This is because the neural network for the CIFAR-10 dataset
has a much larger number of neurons in comparison to the network for
MNIST (see \secref{subsect:setup}). Consequently, there are also many
more inessential neurons.

\paragraph{\textbf{RQ5: Multiple classifiers.}}
As explained in \secref{subsect:P-RAID}, P-\tool is the robust version of
\tool against adaptive adversaries. In this research question, we
investigate the effectiveness of a pool of classifiers versus a single
classifier. In particular, we compare the AUC scores achieved by \tool
and P-\tool, with a pool size of 32 random forests (each with 32
estimators). We trained both tools with the expected attacks (e.g.,
trained with FGSM and tested with FGSM). The results are presented in
\tabref{tab:rq5}.

As shown in the table, the effectiveness of the tools is almost
identical. We mark the three differences in bold; the achieved AUC
scores differ by 0.01 across the two implementations. Consequently,
using the P-\tool version of our approach has no negative impact on
its success in detecting adversarial images. On the contrary, it
offers the theoretical benefit of being more powerful against adaptive
adversaries.

On the other hand, state-of-the-art tools mentioned in RQ3 do not
claim robustness against adaptive adversaries. The original idea
behind LSA is already shown to be
bypassed~\cite{CarliniWagner2017-Bypassing}. For DSA, there is no
specific mechanism against adaptive adversaries, and such
classifier-based detectors are typically found to be weak against
adaptive adversaries~\cite{CarliniWagner2017-Bypassing}.
In our experiments,  we show the weakness of mMutant
by introducing CW-0.95. 

\paragraph{\textbf{RQ6: Different classifier types.}}
We now study the impact of different classifier types and their
parameters on the effectiveness of our approach. In particular, we use
the following classifier types:
\begin{description}
\item[DT:] Decision tree \cite{Quinlan1986}
\item[RF:] Random forest with 32, 64, and 128 estimators \cite{Breiman2001}
\item[AB:] AdaBoost with 32, 64, and 128 estimators \cite{Schapire1999}
\item[KNN:] $k$-nearest neighbors with 3 and 5 neighbors \cite{Omohundro1989, Bentley1975}
\end{description}
We selected these classifiers because they are simple when compared to
huge detection networks \cite{MetzenGenewein2017}, on the other hand,
they are known to be effective in many tasks. Note that RF and AB are
referred to as ensemble methods because they consist of multiple
classifiers, in this case decision trees.

\begin{figure*}[t]
\centering
\scalebox{0.75}{
  \begin{tikzpicture}
    \begin{axis}[
      ybar,
      xticklabels={\textbf{DT},\textbf{RF32},\textbf{RF64},\textbf{RF128},\textbf{AB32},\textbf{AB64},\textbf{AB128},\textbf{KNN3},\textbf{KNN5}},
      xtick=data,
      ylabel={\textbf{AUC}},
      legend cell align=left,
      axis y line*=none,
      axis x line*=bottom,
      width=24cm,
      height=4cm,
      ymin=0.8,
      ymax=1.2,
      ytick distance=0.2,
      bar width=0.4cm,
      area legend,
      every node near coord/.append style={font=\scriptsize},
      nodes near coords*={%
        \pgfmathprintnumber[fixed,fixed zerofill, precision=2]\pgfplotspointmeta},
      legend style={legend pos=north east,font=\small, draw=none, legend columns=-1}
      ]

      \pgfplotstableread{classifier-types-MNIST.dat} \comparison

      \addplot+[ybar, Red!70]
          table[x expr=\coordindex, y=Lall] from \comparison;

      \addplot+[ybar, Blue, pattern color=Blue, pattern=north east lines]
          table[x expr=\coordindex, y=Linf] from \comparison;

      \addplot+[ybar, Green, pattern color=Green, pattern=north west lines]
          table[x expr=\coordindex, y=L2] from \comparison;

      \addplot+[ybar, DarkOrange, pattern color=DarkOrange, pattern=crosshatch]
          table[x expr=\coordindex, y=L0] from \comparison;

      \legend{$\mathbf{L_*}$, $\mathbf{L_\infty}$, $\mathbf{L_2}$, $\mathbf{L_0}$}
    \end{axis}
  \end{tikzpicture}
}
\vspace{-1em}
\caption{AUC scores achieved by \tool on the MNIST dataset when using
  different classifier types.}
\vspace{-0.5em}
\label{fig:rq6MNIST}
\end{figure*}
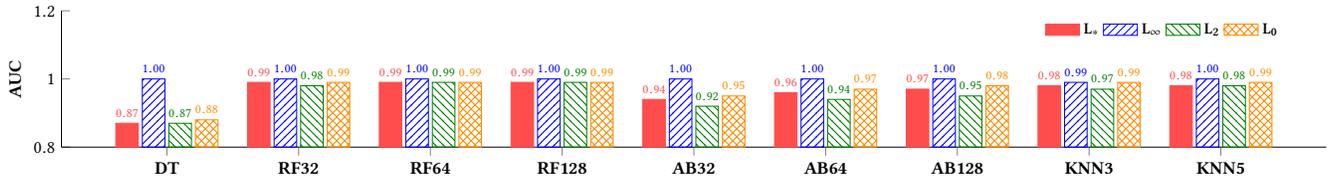

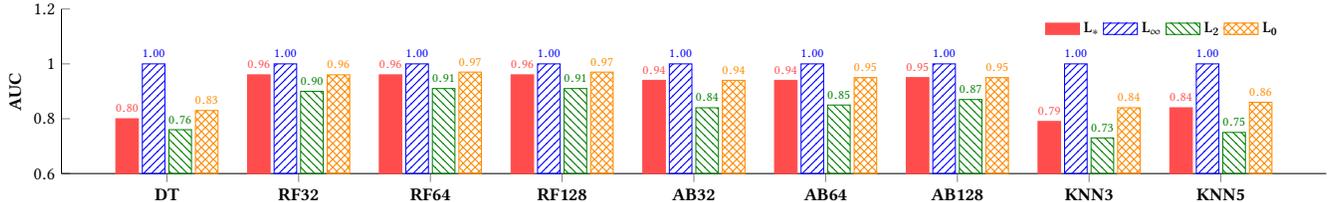
\begin{figure*}[t]
\centering
\scalebox{0.75}{
  \begin{tikzpicture}
    \begin{axis}[
      ybar,
      xticklabels={\textbf{DT},\textbf{RF32},\textbf{RF64},\textbf{RF128},\textbf{AB32},\textbf{AB64},\textbf{AB128},\textbf{KNN3},\textbf{KNN5}},
      xtick=data,
      ylabel={\textbf{AUC}},
      legend cell align=left,
      axis y line*=none,
      axis x line*=bottom,
      width=24cm,
      height=4.5cm,
      ymin=0.6,
      ymax=1.2,
      ytick distance=0.2,
      bar width=0.4cm,
      area legend,
      every node near coord/.append style={font=\scriptsize},
      nodes near coords*={%
        \pgfmathprintnumber[fixed,fixed zerofill, precision=2]\pgfplotspointmeta},
      legend style={legend pos=north east,font=\small, draw=none, legend columns=-1}
      ]

      \pgfplotstableread{classifier-types-CIFAR.dat} \comparison

      \addplot+[ybar, Red!70]
          table[x expr=\coordindex, y=Lall] from \comparison;

      \addplot+[ybar, Blue, pattern color=Blue, pattern=north east lines]
          table[x expr=\coordindex, y=Linf] from \comparison;

      \addplot+[ybar, Green, pattern color=Green, pattern=north west lines]
          table[x expr=\coordindex, y=L2] from \comparison;

      \addplot+[ybar, DarkOrange, pattern color=DarkOrange, pattern=crosshatch]
          table[x expr=\coordindex, y=L0] from \comparison;

      \legend{$\mathbf{L_*}$, $\mathbf{L_\infty}$, $\mathbf{L_2}$, $\mathbf{L_0}$}
    \end{axis}
  \end{tikzpicture}
}
\vspace{-1em}
\caption{AUC scores achieved by \tool on the CIFAR-10 dataset when
  using different classifier types.}
\vspace{-1em}
\label{fig:rq6CIFAR}
\end{figure*}

For this experiment, we train one classifier of each type (not a pool)
with the expected attack norms. We then measure the AUC scores
achieved by \tool when using each of these classifiers. The results
are presented in \figsref{fig:rq6MNIST} and~\ref{fig:rq6CIFAR} for
MNIST and CIFAR-10, respectively. As shown in the bar charts, the
ensemble methods RF and AB are more effective than DT and KNN on the
two datasets. KNN is slightly better than AB for MNIST, but noticeably
worse for CIFAR-10. These results suggest that ensembles of simple
classifiers, such as decision trees, are sufficient for our approach
to be very effective in practice.

Out of the two ensemble methods, RF is more effective than AB. The
number of RF estimators does not have a significant impact on the AUC
scores. Specifically, when using 64 estimators, the AUC scores can
increase by up to 0.01 in comparison to 32 estimators (for
$\text{L}_2$ attacks in both datasets, and for $\text{L}_0$ in
CIFAR-10). There is also no difference between using 64 and 128
estimators. Recall, however, that the number of estimators affects the
training time, and therefore, RF32 can be trained more efficiently.

\begin{table}[t]
\caption{AUC scores achieved by \tool and P-\tool when trained and
  tested on the same attacks.}
\vspace{-1em}
\label{tab:rq5}
\scalebox{0.95}{
\begin{tabular}{c|cc|cc}
\toprule
\multicolumn{1}{l|}{} & \multicolumn{2}{c|}{\textbf{MNIST}}        & \multicolumn{2}{c}{\textbf{CIFAR-10}}     \\
                      & \textbf{\tool} & \textbf{P-\tool} & \textbf{\tool} & \textbf{P-\tool}\\ \midrule
$\mathbf{L_*}$        & 0.99 $\pm$ 0.00 & 0.99 $\pm$ 0.01 & 0.96 $\pm$ 0.00 & 0.96 $\pm$ 0.01\\
$\mathbf{L_\infty}$    & 1.00 $\pm$ 0.00 & 1.00 $\pm$ 0.00 & 1.00 $\pm$ 0.00 & 1.00 $\pm$ 0.00\\
$\mathbf{L_2}$        & 0.98 $\pm$ 0.00 & 0.98 $\pm$ 0.01 & 0.90 $\pm$ 0.00 & 0.90 $\pm$ 0.02\\
\textbf{PGD}          & 1.00 $\pm$ 0.00 & 1.00 $\pm$ 0.00 & 1.00 $\pm$ 0.00 & 1.00 $\pm$ 0.00\\
\textbf{FGSM}         & 1.00 $\pm$ 0.00 & 1.00 $\pm$ 0.00 & 1.00 $\pm$ 0.00 & 1.00 $\pm$ 0.00\\
\textbf{BIM}          & 1.00 $\pm$ 0.00 & 1.00 $\pm$ 0.00 & 1.00 $\pm$ 0.00 & 1.00 $\pm$ 0.00\\
\textbf{DF}           & \bf{0.99 $\pm$ 0.00} & \bf{0.98 $\pm$ 0.00} & 0.91 $\pm$ 0.00 & 0.91 $\pm$ 0.02\\
\textbf{CW}           & 0.98 $\pm$ 0.01 & 0.98 $\pm$ 0.01 & \bf{0.91 $\pm$ 0.01} & \bf{0.90 $\pm$ 0.02}\\
\textbf{JSMA}         & 0.99 $\pm$ 0.00 & 0.99 $\pm$ 0.01 & 0.96 $\pm$ 0.00 & 0.96 $\pm$ 0.01\\ \bottomrule
\end{tabular}}
\vspace{-1.5em}
\end{table}

\subsection{Threats to Validity}
\label{subsect:threats}

We identify the following threats to the validity of our experiments.

\paragraph{\textbf{External validity.}}
External validity ensures that the results of an experimental
evaluation generalize~\cite{SiegmundSiegmund2015}. Our results may not
generalize to other datasets or network
models~\cite{SiegmundSiegmund2015}. However, we use two of the most
popular datasets for evaluating state-of-the-art adversarial-image
detection techniques (e.g., \cite{KimFeldt2019,WangDong2019}) and
borrow the models from one of these
techniques~\cite{KimFeldt2019}. The results may also not generalize to
other attacks although we generate adversarial images using six
well-known attacks across three attack norms.

In our evaluation, we compare \tool with three adversarial-image
detection techniques, but our results may not generalize to
others~\cite{SiegmundSiegmund2015}. To alleviate this threat, we
selected the most recent detection techniques published at top
venues. Moreover, our implementation is open source so that others can
reproduce our results and perform more extensive comparisons.

\paragraph{\textbf{Internal validity.}}
The internal validity~\cite{SiegmundSiegmund2015} of randomized
approaches may be compromised by a potentially biased selection of
random seeds.  We avoid this pitfall by performing all of our
experiments 8 times, each time using a different random seed for the
classifier and a different randomly selected set of neurons. We
report mean results and their standard deviation.


\section{Related Work}
\label{sect:relatedWork}

\paragraph{\textbf{Adversarial robustness.}}
The area of adversarial machine learning has been very active since
the discovery of adversarial examples for neural
networks~\cite{SzegedyZaremba2014}. On one side, more effective
adversarial attacks are being developed. On the other side,
researchers develop new techniques to obtain \emph{robust} neural
networks. Here, we consider two types of robustness techniques: (1) those
that \emph{detect} adversarial examples, and (2)
those that aim to build a network for which it is difficult to
generate adversarial examples, namely \emph{defenses}. According to this
classification, \tool falls into the first category.

\paragraph{\textbf{Adversarial-input detection.}}
Despite these efforts to ensure robustness of neural networks, new
attacks often evade existing defense or detection techniques. On the
detection side, Grosse et al., Gong et al., and Metzen et
al.~\cite{GrosseManoharan2017,GongWang2017,MetzenGenewein2017} train a
secondary classifier for detecting adversarial examples; in this
aspect, these techniques are similar to ours. From these three, the
approach presented by Metzen et al.~\cite{MetzenGenewein2017} is the
closest to our work. However, unlike their technique, we propose a
much simpler methodology and an extension (i.e., in P-\tool) to handle
adaptive adversaries. Hendrycks and Gimpel, Bhagoji et al., and Li and
Li~\cite{HendrycksGimpel2017,BhagojiCullina2017,LiLi2017} propose
methods based on Principal Component Analysis (PCA). Feinman et
al.~\cite{FeinmanCurtin2017} introduce two techniques, called
kernel-density estimation and Bayesian neural-network
uncertainty. Carlini and Wagner~\cite{CarliniWagner2017-Bypassing},
however, find all of the aforementioned techniques ineffective for
providing a thorough detection mechanism.

While there are other works that claim to be effective in detecting
adversarial inputs, they have already been surpassed or bypassed by
later work. For example, work on statistical
detection~\cite{RothKilcher2019} can be
bypassed~\cite{HosseiniKannan2019}. Similarly, feature
squeezing~\cite{XuEvans2018} can be defeated~\cite{SharmaChen2018}, and
both methods by Zantedeschi et al.~\cite{ZantedeschiNicolae2017} and
MagNet~\cite{MengChen2017} are shown not to be
robust~\cite{CarliniWagner2017-MagNet,LuChen2018}.

Even more recent detection techniques include work that specializes on
adaptive adversaries~\cite{HuYu2019}, focuses on black-box
attacks~\cite{ChenCarlini2019}, and makes use of the SHAP
explainability technique~\cite{FidelBitton2019}. Lastly, two detection
techniques~\cite{KimFeldt2019,WangDong2019} have recently been
proposed by the software-engineering community. We provide a detailed
description of these in the previous section, where we compare them
with \tool.

\paragraph{\textbf{Defenses against adversarial inputs.}}
On the defense side, adversarial retraining was found to be
effective~\cite{MadryMakelov2018}. Despite this, many
defenses~\cite{PapernotMcDaniel2016-Distillation,NayebiGanguli2017}
are circumvented by other
work~\cite{AthalyeCarlini2018-Gradients,HeWei2017,CarliniWagner2016}. Generally,
one should follow recent guidelines for properly evaluating the
robustness of a neural network~\cite{CarliniAthalye2019}.

\paragraph{\textbf{Neural-network testing.}}
The increasing popularity of deep-learning systems and their
vulnerability to adversarial examples also motivated research into
testing techniques for neural networks.  Existing research in the area
adapts testing methodologies from traditional software
engineering. DeepXplore~\cite{PeiCao2017} is the first to introduce a
specialized coverage criterion (i.e., neuron coverage) for neural
networks. Subsequently, new such coverage criteria have been
proposed. For example, DeepGauge~\cite{MaJuefeiXu2018} refines neuron
coverage, DeepCover~\cite{SunHuang2018} adapts MC/DC from traditional
software testing, and DeepCT~\cite{MaZhang2018} investigates
combinatorial testing. However, more recent work argues that there is
limited correlation between coverage and robustness of neural
networks~\cite{DongZhang2019}.

Eniser et al.~\cite{EniserGerasimou2019} make use of
fault-localization metrics for finding neurons that can be exploited
to fool networks. Sun et al.~\cite{SunWu2018} adapt concolic
testing~\cite{GodefroidKlarlund2005,CadarEngler2005,SenMarinov2005}
for generating test inputs to neural networks. Other approaches make
use of coverage-guided fuzzing for input
generation~\cite{OdenaOlsson2019,XieMa2019,DemirEniser2019}. DeepTest~\cite{TianPei2018}
and DeepRoad~\cite{ZhangZhang2018} focus on generating test inputs for
deep-learning-based autonomous driving systems. Unlike the above
testing techniques, \tool proposes an \emph{online} detection
technique for adversarial examples.

\section{Conclusion}
\label{sect:conclusion}

In this work, we propose a novel mechanism, namely \tool, for detecting adversarial
examples in neural networks. In addition to this, we introduce P-\tool, which is designed
to be robust even against adaptive adversaries.  We extensively evaluate our technique and
show its effectiveness, for instance by achieving a 90\% AUC score against the strongest
attacks (i.e., CW, DF) and perfect detection against weaker adversaries (i.e., PGD, BIM,
FGSM). The comparison with three state-of-the-art detection techniques shows that \tool is
both more stable and more effective. In the future, we plan to test our tool on other
threat models, larger neural networks, and different tasks, such as natural language
processing.



\newpage

\bibliographystyle{ACM-Reference-Format}
\bibliography{paper}


\begin{thebibliography}{64}


\ifx \showCODEN    \undefined \def \showCODEN     #1{\unskip}     \fi
\ifx \showDOI      \undefined \def \showDOI       #1{#1}\fi
\ifx \showISBNx    \undefined \def \showISBNx     #1{\unskip}     \fi
\ifx \showISBNxiii \undefined \def \showISBNxiii  #1{\unskip}     \fi
\ifx \showISSN     \undefined \def \showISSN      #1{\unskip}     \fi
\ifx \showLCCN     \undefined \def \showLCCN      #1{\unskip}     \fi
\ifx \shownote     \undefined \def \shownote      #1{#1}          \fi
\ifx \showarticletitle \undefined \def \showarticletitle #1{#1}   \fi
\ifx \showURL      \undefined \def \showURL       {\relax}        \fi
\providecommand\bibfield[2]{#2}
\providecommand\bibinfo[2]{#2}
\providecommand\natexlab[1]{#1}
\providecommand\showeprint[2][]{arXiv:#2}

\bibitem[\protect\citeauthoryear{??}{Ker}{[n.d.]}]%
        {Keras}
 \bibinfo{year}{[n.d.]}\natexlab{}.
\newblock \bibinfo{title}{{Keras}: {T}he {Python} Deep Learning Library}.
\newblock
\newblock
\newblock
\shownote{\url{https://keras.io}.}


\bibitem[\protect\citeauthoryear{??}{MNI}{[n.d.]}]%
        {MNIST}
 \bibinfo{year}{[n.d.]}\natexlab{}.
\newblock \bibinfo{title}{The {MNIST} Database of Handwritten Digits}.
\newblock
\newblock
\newblock
\shownote{\url{http://yann.lecun.com/exdb/mnist}.}


\bibitem[\protect\citeauthoryear{??}{Ten}{[n.d.]}]%
        {Tensorflow}
 \bibinfo{year}{[n.d.]}\natexlab{}.
\newblock \bibinfo{title}{{TensorFlow}: {L}arge-Scale Machine Learning on
  Heterogeneous Systems}.
\newblock
\newblock
\newblock
\shownote{\url{https://www.tensorflow.org}.}


\bibitem[\protect\citeauthoryear{Athalye, Carlini, and Wagner}{Athalye
  et~al\mbox{.}}{2018}]%
        {AthalyeCarlini2018-Gradients}
\bibfield{author}{\bibinfo{person}{Anish Athalye}, \bibinfo{person}{Nicholas
  Carlini}, {and} \bibinfo{person}{David~A. Wagner}.}
  \bibinfo{year}{2018}\natexlab{}.
\newblock \showarticletitle{Obfuscated Gradients Give a False Sense of
  Security: {C}ircumventing Defenses to Adversarial Examples}. In
  \bibinfo{booktitle}{\emph{ICML}} \emph{(\bibinfo{series}{PMLR})},
  Vol.~\bibinfo{volume}{80}. \bibinfo{publisher}{PMLR},
  \bibinfo{pages}{274--283}.
\newblock


\bibitem[\protect\citeauthoryear{Bentley}{Bentley}{1975}]%
        {Bentley1975}
\bibfield{author}{\bibinfo{person}{Jon~Louis Bentley}.}
  \bibinfo{year}{1975}\natexlab{}.
\newblock \showarticletitle{Multidimensional Binary Search Trees Used for
  Associative Searching}.
\newblock \bibinfo{journal}{\emph{CACM}}  \bibinfo{volume}{18}
  (\bibinfo{year}{1975}), \bibinfo{pages}{509--517}.
\newblock
Issue 9.


\bibitem[\protect\citeauthoryear{Bhagoji, Cullina, and Mittal}{Bhagoji
  et~al\mbox{.}}{2017}]%
        {BhagojiCullina2017}
\bibfield{author}{\bibinfo{person}{Arjun~Nitin Bhagoji},
  \bibinfo{person}{Daniel Cullina}, {and} \bibinfo{person}{Prateek Mittal}.}
  \bibinfo{year}{2017}\natexlab{}.
\newblock \showarticletitle{Dimensionality Reduction as a Defense Against
  Evasion Attacks on Machine Learning Classifiers}.
\newblock \bibinfo{journal}{\emph{CoRR}}  \bibinfo{volume}{abs/1704.02654}
  (\bibinfo{year}{2017}).
\newblock


\bibitem[\protect\citeauthoryear{Breiman}{Breiman}{2001}]%
        {Breiman2001}
\bibfield{author}{\bibinfo{person}{Leo Breiman}.}
  \bibinfo{year}{2001}\natexlab{}.
\newblock \showarticletitle{Random Forests}.
\newblock \bibinfo{journal}{\emph{Machine Learning}}  \bibinfo{volume}{45}
  (\bibinfo{year}{2001}), \bibinfo{pages}{5--32}.
\newblock
Issue 1.


\bibitem[\protect\citeauthoryear{Cadar and Engler}{Cadar and Engler}{2005}]%
        {CadarEngler2005}
\bibfield{author}{\bibinfo{person}{Cristian Cadar} {and}
  \bibinfo{person}{Dawson~R. Engler}.} \bibinfo{year}{2005}\natexlab{}.
\newblock \showarticletitle{Execution Generated Test Cases: {H}ow to Make
  Systems Code Crash Itself}. In \bibinfo{booktitle}{\emph{SPIN}}
  \emph{(\bibinfo{series}{LNCS})}, Vol.~\bibinfo{volume}{3639}.
  \bibinfo{publisher}{Springer}, \bibinfo{pages}{2--23}.
\newblock


\bibitem[\protect\citeauthoryear{Carlini, Athalye, Papernot, Brendel, Rauber,
  Tsipras, Goodfellow, Madry, and Kurakin}{Carlini et~al\mbox{.}}{2019}]%
        {CarliniAthalye2019}
\bibfield{author}{\bibinfo{person}{Nicholas Carlini}, \bibinfo{person}{Anish
  Athalye}, \bibinfo{person}{Nicolas Papernot}, \bibinfo{person}{Wieland
  Brendel}, \bibinfo{person}{Jonas Rauber}, \bibinfo{person}{Dimitris Tsipras},
  \bibinfo{person}{Ian~J. Goodfellow}, \bibinfo{person}{Aleksander Madry},
  {and} \bibinfo{person}{Alexey Kurakin}.} \bibinfo{year}{2019}\natexlab{}.
\newblock \showarticletitle{On Evaluating Adversarial Robustness}.
\newblock \bibinfo{journal}{\emph{CoRR}}  \bibinfo{volume}{abs/1902.06705}
  (\bibinfo{year}{2019}).
\newblock


\bibitem[\protect\citeauthoryear{Carlini and Wagner}{Carlini and
  Wagner}{2016}]%
        {CarliniWagner2016}
\bibfield{author}{\bibinfo{person}{Nicholas Carlini} {and}
  \bibinfo{person}{David~A. Wagner}.} \bibinfo{year}{2016}\natexlab{}.
\newblock \showarticletitle{Defensive Distillation is Not Robust to Adversarial
  Examples}.
\newblock \bibinfo{journal}{\emph{CoRR}}  \bibinfo{volume}{abs/1607.04311}
  (\bibinfo{year}{2016}).
\newblock


\bibitem[\protect\citeauthoryear{Carlini and Wagner}{Carlini and
  Wagner}{2017a}]%
        {CarliniWagner2017-Bypassing}
\bibfield{author}{\bibinfo{person}{Nicholas Carlini} {and}
  \bibinfo{person}{David~A. Wagner}.} \bibinfo{year}{2017}\natexlab{a}.
\newblock \showarticletitle{Adversarial Examples Are Not Easily Detected:
  {B}ypassing Ten Detection Methods}. In \bibinfo{booktitle}{\emph{AISec@CCS}}.
  \bibinfo{publisher}{ACM}, \bibinfo{pages}{3--14}.
\newblock


\bibitem[\protect\citeauthoryear{Carlini and Wagner}{Carlini and
  Wagner}{2017b}]%
        {CarliniWagner2017-MagNet}
\bibfield{author}{\bibinfo{person}{Nicholas Carlini} {and}
  \bibinfo{person}{David~A. Wagner}.} \bibinfo{year}{2017}\natexlab{b}.
\newblock \showarticletitle{{MagNet} and ``Efficient Defenses Against
  Adversarial Attacks'' Are Not Robust to Adversarial Examples}.
\newblock \bibinfo{journal}{\emph{CoRR}}  \bibinfo{volume}{abs/1711.08478}
  (\bibinfo{year}{2017}).
\newblock


\bibitem[\protect\citeauthoryear{Carlini and Wagner}{Carlini and
  Wagner}{2017c}]%
        {CarliniWagner2017-Robustness}
\bibfield{author}{\bibinfo{person}{Nicholas Carlini} {and}
  \bibinfo{person}{David~A. Wagner}.} \bibinfo{year}{2017}\natexlab{c}.
\newblock \showarticletitle{Towards Evaluating the Robustness of Neural
  Networks}. In \bibinfo{booktitle}{\emph{S\&P}}. \bibinfo{publisher}{IEEE
  Computer Society}, \bibinfo{pages}{39--57}.
\newblock


\bibitem[\protect\citeauthoryear{Chen, Carlini, and Wagner}{Chen
  et~al\mbox{.}}{2019}]%
        {ChenCarlini2019}
\bibfield{author}{\bibinfo{person}{Steven Chen}, \bibinfo{person}{Nicholas
  Carlini}, {and} \bibinfo{person}{David~A. Wagner}.}
  \bibinfo{year}{2019}\natexlab{}.
\newblock \showarticletitle{Stateful Detection of Black-Box Adversarial
  Attacks}.
\newblock \bibinfo{journal}{\emph{CoRR}}  \bibinfo{volume}{abs/1907.05587}
  (\bibinfo{year}{2019}).
\newblock


\bibitem[\protect\citeauthoryear{Demir, Eniser, and Sen}{Demir
  et~al\mbox{.}}{2019}]%
        {DemirEniser2019}
\bibfield{author}{\bibinfo{person}{Samet Demir}, \bibinfo{person}{Hasan~Ferit
  Eniser}, {and} \bibinfo{person}{Alper Sen}.} \bibinfo{year}{2019}\natexlab{}.
\newblock \showarticletitle{{DeepSmartFuzzer}: {R}eward Guided Test Generation
  For Deep Learning}.
\newblock \bibinfo{journal}{\emph{CoRR}}  \bibinfo{volume}{abs/1911.10621}
  (\bibinfo{year}{2019}).
\newblock


\bibitem[\protect\citeauthoryear{Dong, Zhang, Wang, Liu, Sun, Hao, Wang, Wang,
  Dong, and Ting}{Dong et~al\mbox{.}}{2019}]%
        {DongZhang2019}
\bibfield{author}{\bibinfo{person}{Yizhen Dong}, \bibinfo{person}{Peixin
  Zhang}, \bibinfo{person}{Jingyi Wang}, \bibinfo{person}{Shuang Liu},
  \bibinfo{person}{Jun Sun}, \bibinfo{person}{Jianye Hao},
  \bibinfo{person}{Xinyu Wang}, \bibinfo{person}{Li Wang},
  \bibinfo{person}{Jin~Song Dong}, {and} \bibinfo{person}{Dai Ting}.}
  \bibinfo{year}{2019}\natexlab{}.
\newblock \showarticletitle{There is Limited Correlation between Coverage and
  Robustness for Deep Neural Networks}.
\newblock \bibinfo{journal}{\emph{CoRR}}  \bibinfo{volume}{abs/1911.05904}
  (\bibinfo{year}{2019}).
\newblock


\bibitem[\protect\citeauthoryear{Eniser, Gerasimou, and Sen}{Eniser
  et~al\mbox{.}}{2019}]%
        {EniserGerasimou2019}
\bibfield{author}{\bibinfo{person}{Hasan~Ferit Eniser}, \bibinfo{person}{Simos
  Gerasimou}, {and} \bibinfo{person}{Alper Sen}.}
  \bibinfo{year}{2019}\natexlab{}.
\newblock \showarticletitle{{DeepFault}: {F}ault Localization for Deep Neural
  Networks}. In \bibinfo{booktitle}{\emph{FASE}}
  \emph{(\bibinfo{series}{LNCS})}, Vol.~\bibinfo{volume}{11424}.
  \bibinfo{publisher}{Springer}, \bibinfo{pages}{171--191}.
\newblock


\bibitem[\protect\citeauthoryear{Eykholt, Evtimov, Fernandes, Li, Rahmati,
  Xiao, Prakash, Kohno, and Song}{Eykholt et~al\mbox{.}}{2018}]%
        {EykholtEvtimov2018}
\bibfield{author}{\bibinfo{person}{Kevin Eykholt}, \bibinfo{person}{Ivan
  Evtimov}, \bibinfo{person}{Earlence Fernandes}, \bibinfo{person}{Bo Li},
  \bibinfo{person}{Amir Rahmati}, \bibinfo{person}{Chaowei Xiao},
  \bibinfo{person}{Atul Prakash}, \bibinfo{person}{Tadayoshi Kohno}, {and}
  \bibinfo{person}{Dawn Song}.} \bibinfo{year}{2018}\natexlab{}.
\newblock \showarticletitle{Robust Physical-World Attacks on Deep Learning
  Visual Classification}. In \bibinfo{booktitle}{\emph{CVPR}}.
  \bibinfo{publisher}{IEEE Computer Society}, \bibinfo{pages}{1625--1634}.
\newblock


\bibitem[\protect\citeauthoryear{Feinman, Curtin, Shintre, and Gardner}{Feinman
  et~al\mbox{.}}{2017}]%
        {FeinmanCurtin2017}
\bibfield{author}{\bibinfo{person}{Reuben Feinman}, \bibinfo{person}{Ryan~R.
  Curtin}, \bibinfo{person}{Saurabh Shintre}, {and} \bibinfo{person}{Andrew~B.
  Gardner}.} \bibinfo{year}{2017}\natexlab{}.
\newblock \showarticletitle{Detecting Adversarial Samples from Artifacts}.
\newblock \bibinfo{journal}{\emph{CoRR}}  \bibinfo{volume}{abs/1703.00410}
  (\bibinfo{year}{2017}).
\newblock


\bibitem[\protect\citeauthoryear{Fidel, Bitton, and Shabtai}{Fidel
  et~al\mbox{.}}{2019}]%
        {FidelBitton2019}
\bibfield{author}{\bibinfo{person}{Gil Fidel}, \bibinfo{person}{Ron Bitton},
  {and} \bibinfo{person}{Asaf Shabtai}.} \bibinfo{year}{2019}\natexlab{}.
\newblock \showarticletitle{When Explainability Meets Adversarial Learning:
  {D}etecting Adversarial Examples Using {SHAP} Signatures}.
\newblock \bibinfo{journal}{\emph{CoRR}}  \bibinfo{volume}{abs/1909.03418}
  (\bibinfo{year}{2019}).
\newblock


\bibitem[\protect\citeauthoryear{Godefroid, Klarlund, and Sen}{Godefroid
  et~al\mbox{.}}{2005}]%
        {GodefroidKlarlund2005}
\bibfield{author}{\bibinfo{person}{Patrice Godefroid}, \bibinfo{person}{Nils
  Klarlund}, {and} \bibinfo{person}{Koushik Sen}.}
  \bibinfo{year}{2005}\natexlab{}.
\newblock \showarticletitle{{DART}: {D}irected Automated Random Testing}. In
  \bibinfo{booktitle}{\emph{PLDI}}. \bibinfo{publisher}{ACM},
  \bibinfo{pages}{213--223}.
\newblock


\bibitem[\protect\citeauthoryear{Gong, Wang, and Ku}{Gong
  et~al\mbox{.}}{2017}]%
        {GongWang2017}
\bibfield{author}{\bibinfo{person}{Zhitao Gong}, \bibinfo{person}{Wenlu Wang},
  {and} \bibinfo{person}{Wei{-}Shinn Ku}.} \bibinfo{year}{2017}\natexlab{}.
\newblock \showarticletitle{Adversarial and Clean Data Are Not Twins}.
\newblock \bibinfo{journal}{\emph{CoRR}}  \bibinfo{volume}{abs/1704.04960}
  (\bibinfo{year}{2017}).
\newblock


\bibitem[\protect\citeauthoryear{Goodfellow, Shlens, and Szegedy}{Goodfellow
  et~al\mbox{.}}{2015}]%
        {GoodfellowShlens2015}
\bibfield{author}{\bibinfo{person}{Ian~J. Goodfellow},
  \bibinfo{person}{Jonathon Shlens}, {and} \bibinfo{person}{Christian
  Szegedy}.} \bibinfo{year}{2015}\natexlab{}.
\newblock \showarticletitle{Explaining and Harnessing Adversarial Examples}. In
  \bibinfo{booktitle}{\emph{ICLR}}.
\newblock


\bibitem[\protect\citeauthoryear{Grosse, Manoharan, Papernot, Backes, and
  McDaniel}{Grosse et~al\mbox{.}}{2017}]%
        {GrosseManoharan2017}
\bibfield{author}{\bibinfo{person}{Kathrin Grosse}, \bibinfo{person}{Praveen
  Manoharan}, \bibinfo{person}{Nicolas Papernot}, \bibinfo{person}{Michael
  Backes}, {and} \bibinfo{person}{Patrick~D. McDaniel}.}
  \bibinfo{year}{2017}\natexlab{}.
\newblock \showarticletitle{On the (Statistical) Detection of Adversarial
  Examples}.
\newblock \bibinfo{journal}{\emph{CoRR}}  \bibinfo{volume}{abs/1702.06280}
  (\bibinfo{year}{2017}).
\newblock


\bibitem[\protect\citeauthoryear{He, Wei, Chen, Carlini, and Song}{He
  et~al\mbox{.}}{2017}]%
        {HeWei2017}
\bibfield{author}{\bibinfo{person}{Warren He}, \bibinfo{person}{James Wei},
  \bibinfo{person}{Xinyun Chen}, \bibinfo{person}{Nicholas Carlini}, {and}
  \bibinfo{person}{Dawn Song}.} \bibinfo{year}{2017}\natexlab{}.
\newblock \showarticletitle{Adversarial Example Defense: {E}nsembles of Weak
  Defenses Are Not Strong}. In \bibinfo{booktitle}{\emph{WOOT}}.
  \bibinfo{publisher}{USENIX}.
\newblock


\bibitem[\protect\citeauthoryear{Hendrycks and Gimpel}{Hendrycks and
  Gimpel}{2017}]%
        {HendrycksGimpel2017}
\bibfield{author}{\bibinfo{person}{Dan Hendrycks} {and} \bibinfo{person}{Kevin
  Gimpel}.} \bibinfo{year}{2017}\natexlab{}.
\newblock \showarticletitle{Early Methods for Detecting Adversarial Images}. In
  \bibinfo{booktitle}{\emph{ICLR (Workshop)}}.
  \bibinfo{publisher}{OpenReview.net}.
\newblock


\bibitem[\protect\citeauthoryear{Hosseini, Kannan, and Poovendran}{Hosseini
  et~al\mbox{.}}{2019}]%
        {HosseiniKannan2019}
\bibfield{author}{\bibinfo{person}{Hossein Hosseini}, \bibinfo{person}{Sreeram
  Kannan}, {and} \bibinfo{person}{Radha Poovendran}.}
  \bibinfo{year}{2019}\natexlab{}.
\newblock \showarticletitle{Are Odds Really Odd? Bypassing Statistical
  Detection of Adversarial Examples}.
\newblock \bibinfo{journal}{\emph{CoRR}}  \bibinfo{volume}{abs/1907.12138}
  (\bibinfo{year}{2019}).
\newblock


\bibitem[\protect\citeauthoryear{Hu, Yu, Guo, Chao, and Weinberger}{Hu
  et~al\mbox{.}}{2019}]%
        {HuYu2019}
\bibfield{author}{\bibinfo{person}{Shengyuan Hu}, \bibinfo{person}{Tao Yu},
  \bibinfo{person}{Chuan Guo}, \bibinfo{person}{Wei-Lun Chao}, {and}
  \bibinfo{person}{Kilian~Q. Weinberger}.} \bibinfo{year}{2019}\natexlab{}.
\newblock \showarticletitle{A New Defense Against Adversarial Images: {T}urning
  a Weakness into a Strength}. In \bibinfo{booktitle}{\emph{NeurIPS}}.
  \bibinfo{pages}{1633--1644}.
\newblock


\bibitem[\protect\citeauthoryear{Kim, Feldt, and Yoo}{Kim
  et~al\mbox{.}}{2019}]%
        {KimFeldt2019}
\bibfield{author}{\bibinfo{person}{Jinhan Kim}, \bibinfo{person}{Robert Feldt},
  {and} \bibinfo{person}{Shin Yoo}.} \bibinfo{year}{2019}\natexlab{}.
\newblock \showarticletitle{Guiding Deep Learning System Testing Using Surprise
  Adequacy}. In \bibinfo{booktitle}{\emph{ICSE}}. \bibinfo{publisher}{IEEE
  Computer Society/ACM}, \bibinfo{pages}{1039--1049}.
\newblock


\bibitem[\protect\citeauthoryear{Krizhevsky}{Krizhevsky}{2009}]%
        {Krizhevsky2008}
\bibfield{author}{\bibinfo{person}{Alex Krizhevsky}.}
  \bibinfo{year}{2009}\natexlab{}.
\newblock \bibinfo{booktitle}{\emph{Learning Multiple Layers of Features from
  Tiny Images}}.
\newblock \bibinfo{type}{{T}echnical {R}eport}.
  \bibinfo{institution}{University of Toronto}.
\newblock


\bibitem[\protect\citeauthoryear{Kurakin, Goodfellow, and Bengio}{Kurakin
  et~al\mbox{.}}{2017}]%
        {KurakinGoodfellow2017}
\bibfield{author}{\bibinfo{person}{Alexey Kurakin}, \bibinfo{person}{Ian~J.
  Goodfellow}, {and} \bibinfo{person}{Samy Bengio}.}
  \bibinfo{year}{2017}\natexlab{}.
\newblock \showarticletitle{Adversarial Examples in the Physical World}. In
  \bibinfo{booktitle}{\emph{ICLR}}. \bibinfo{publisher}{OpenReview.net}.
\newblock


\bibitem[\protect\citeauthoryear{Li and Li}{Li and Li}{2017}]%
        {LiLi2017}
\bibfield{author}{\bibinfo{person}{Xin Li} {and} \bibinfo{person}{Fuxin Li}.}
  \bibinfo{year}{2017}\natexlab{}.
\newblock \showarticletitle{Adversarial Examples Detection in Deep Networks
  with Convolutional Filter Statistics}. In \bibinfo{booktitle}{\emph{ICCV)}}.
  \bibinfo{publisher}{IEEE Computer Society}, \bibinfo{pages}{5775--5783}.
\newblock


\bibitem[\protect\citeauthoryear{Lu, Chen, Chen, and Yu}{Lu
  et~al\mbox{.}}{2018}]%
        {LuChen2018}
\bibfield{author}{\bibinfo{person}{Pei{-}Hsuan Lu}, \bibinfo{person}{Pin{-}Yu
  Chen}, \bibinfo{person}{Kang{-}Cheng Chen}, {and} \bibinfo{person}{Chia{-}Mu
  Yu}.} \bibinfo{year}{2018}\natexlab{}.
\newblock \showarticletitle{On the Limitation of {MagNet} Defense Against
  {L1}-Based Adversarial Examples}. In \bibinfo{booktitle}{\emph{DSN
  Workshops}}. \bibinfo{publisher}{IEEE Computer Society},
  \bibinfo{pages}{200--214}.
\newblock


\bibitem[\protect\citeauthoryear{Ma, Juefei{-}Xu, Zhang, Sun, Xue, Li, Chen,
  Su, Li, Liu, Zhao, and Wang}{Ma et~al\mbox{.}}{2018a}]%
        {MaJuefeiXu2018}
\bibfield{author}{\bibinfo{person}{Lei Ma}, \bibinfo{person}{Felix
  Juefei{-}Xu}, \bibinfo{person}{Fuyuan Zhang}, \bibinfo{person}{Jiyuan Sun},
  \bibinfo{person}{Minhui Xue}, \bibinfo{person}{Bo Li},
  \bibinfo{person}{Chunyang Chen}, \bibinfo{person}{Ting Su},
  \bibinfo{person}{Li Li}, \bibinfo{person}{Yang Liu}, \bibinfo{person}{Jianjun
  Zhao}, {and} \bibinfo{person}{Yadong Wang}.}
  \bibinfo{year}{2018}\natexlab{a}.
\newblock \showarticletitle{{DeepGauge}: {M}ulti-Granularity Testing Criteria
  for Deep Learning Systems}. In \bibinfo{booktitle}{\emph{ASE}}.
  \bibinfo{publisher}{ACM}, \bibinfo{pages}{120--131}.
\newblock


\bibitem[\protect\citeauthoryear{Ma, Zhang, Xue, Li, Liu, Zhao, and Wang}{Ma
  et~al\mbox{.}}{2018b}]%
        {MaZhang2018}
\bibfield{author}{\bibinfo{person}{Lei Ma}, \bibinfo{person}{Fuyuan Zhang},
  \bibinfo{person}{Minhui Xue}, \bibinfo{person}{Bo Li}, \bibinfo{person}{Yang
  Liu}, \bibinfo{person}{Jianjun Zhao}, {and} \bibinfo{person}{Yadong Wang}.}
  \bibinfo{year}{2018}\natexlab{b}.
\newblock \showarticletitle{Combinatorial Testing for Deep Learning Systems}.
\newblock \bibinfo{journal}{\emph{CoRR}}  \bibinfo{volume}{abs/1806.07723}
  (\bibinfo{year}{2018}).
\newblock


\bibitem[\protect\citeauthoryear{Madry, Makelov, Schmidt, Tsipras, and
  Vladu}{Madry et~al\mbox{.}}{2018}]%
        {MadryMakelov2018}
\bibfield{author}{\bibinfo{person}{Aleksander Madry},
  \bibinfo{person}{Aleksandar Makelov}, \bibinfo{person}{Ludwig Schmidt},
  \bibinfo{person}{Dimitris Tsipras}, {and} \bibinfo{person}{Adrian Vladu}.}
  \bibinfo{year}{2018}\natexlab{}.
\newblock \showarticletitle{Towards Deep Learning Models Resistant to
  Adversarial Attacks}. In \bibinfo{booktitle}{\emph{ICLR}}.
  \bibinfo{publisher}{OpenReview.net}.
\newblock


\bibitem[\protect\citeauthoryear{Meng and Chen}{Meng and Chen}{2017}]%
        {MengChen2017}
\bibfield{author}{\bibinfo{person}{Dongyu Meng} {and} \bibinfo{person}{Hao
  Chen}.} \bibinfo{year}{2017}\natexlab{}.
\newblock \showarticletitle{{MagNet}: {A} Two-Pronged Defense Against
  Adversarial Examples}. In \bibinfo{booktitle}{\emph{CCS}}.
  \bibinfo{publisher}{ACM}, \bibinfo{pages}{135–--147}.
\newblock


\bibitem[\protect\citeauthoryear{Metzen, Genewein, Fischer, and
  Bischoff}{Metzen et~al\mbox{.}}{2017}]%
        {MetzenGenewein2017}
\bibfield{author}{\bibinfo{person}{Jan~Hendrik Metzen}, \bibinfo{person}{Tim
  Genewein}, \bibinfo{person}{Volker Fischer}, {and} \bibinfo{person}{Bastian
  Bischoff}.} \bibinfo{year}{2017}\natexlab{}.
\newblock \showarticletitle{On Detecting Adversarial Perturbations}.
\newblock \bibinfo{journal}{\emph{CoRR}}  \bibinfo{volume}{abs/1702.04267}
  (\bibinfo{year}{2017}).
\newblock


\bibitem[\protect\citeauthoryear{Moosavi{-}Dezfooli, Fawzi, and
  Frossard}{Moosavi{-}Dezfooli et~al\mbox{.}}{2016}]%
        {MoosaviDezfooliFawzi2016}
\bibfield{author}{\bibinfo{person}{Seyed{-}Mohsen Moosavi{-}Dezfooli},
  \bibinfo{person}{Alhussein Fawzi}, {and} \bibinfo{person}{Pascal Frossard}.}
  \bibinfo{year}{2016}\natexlab{}.
\newblock \showarticletitle{{DeepFool}: {A} Simple and Accurate Method to Fool
  Deep Neural Networks}. In \bibinfo{booktitle}{\emph{CVPR}}.
  \bibinfo{publisher}{IEEE Computer Society}, \bibinfo{pages}{2574--2582}.
\newblock


\bibitem[\protect\citeauthoryear{Nayebi and Ganguli}{Nayebi and
  Ganguli}{2017}]%
        {NayebiGanguli2017}
\bibfield{author}{\bibinfo{person}{Aran Nayebi} {and} \bibinfo{person}{Surya
  Ganguli}.} \bibinfo{year}{2017}\natexlab{}.
\newblock \showarticletitle{Biologically Inspired Protection of Deep Networks
  from Adversarial Attacks}.
\newblock \bibinfo{journal}{\emph{CoRR}}  \bibinfo{volume}{abs/1703.09202}
  (\bibinfo{year}{2017}).
\newblock


\bibitem[\protect\citeauthoryear{Nicolae, Sinn, Minh, Rawat, Wistuba,
  Zantedeschi, Molloy, and Edwards}{Nicolae et~al\mbox{.}}{2018}]%
        {NicolaeSinn2018}
\bibfield{author}{\bibinfo{person}{Maria{-}Irina Nicolae},
  \bibinfo{person}{Mathieu Sinn}, \bibinfo{person}{Tran~Ngoc Minh},
  \bibinfo{person}{Ambrish Rawat}, \bibinfo{person}{Martin Wistuba},
  \bibinfo{person}{Valentina Zantedeschi}, \bibinfo{person}{Ian~M. Molloy},
  {and} \bibinfo{person}{Benjamin Edwards}.} \bibinfo{year}{2018}\natexlab{}.
\newblock \showarticletitle{Adversarial Robustness Toolbox {v0.2.2}}.
\newblock \bibinfo{journal}{\emph{CoRR}}  \bibinfo{volume}{abs/1807.01069}
  (\bibinfo{year}{2018}).
\newblock


\bibitem[\protect\citeauthoryear{Odena, Olsson, Andersen, and Goodfellow}{Odena
  et~al\mbox{.}}{2019}]%
        {OdenaOlsson2019}
\bibfield{author}{\bibinfo{person}{Augustus Odena}, \bibinfo{person}{Catherine
  Olsson}, \bibinfo{person}{David Andersen}, {and} \bibinfo{person}{Ian~J.
  Goodfellow}.} \bibinfo{year}{2019}\natexlab{}.
\newblock \showarticletitle{{TensorFuzz}: {D}ebugging Neural Networks with
  Coverage-Guided Fuzzing}. In \bibinfo{booktitle}{\emph{ICML}}
  \emph{(\bibinfo{series}{PMLR})}, Vol.~\bibinfo{volume}{97}.
  \bibinfo{publisher}{PMLR}, \bibinfo{pages}{4901--4911}.
\newblock


\bibitem[\protect\citeauthoryear{Omohundro}{Omohundro}{1989}]%
        {Omohundro1989}
\bibfield{author}{\bibinfo{person}{Stephen~M. Omohundro}.}
  \bibinfo{year}{1989}\natexlab{}.
\newblock \bibinfo{booktitle}{\emph{Five Balltree Construction Algorithms}}.
\newblock \bibinfo{type}{{T}echnical {R}eport}. \bibinfo{institution}{ICSI}.
\newblock


\bibitem[\protect\citeauthoryear{Papernot, McDaniel, Goodfellow, Jha, Celik,
  and Swami}{Papernot et~al\mbox{.}}{2017}]%
        {PapernotMcDaniel2017}
\bibfield{author}{\bibinfo{person}{Nicolas Papernot},
  \bibinfo{person}{Patrick~D. McDaniel}, \bibinfo{person}{Ian~J. Goodfellow},
  \bibinfo{person}{Somesh Jha}, \bibinfo{person}{Z.~Berkay Celik}, {and}
  \bibinfo{person}{Ananthram Swami}.} \bibinfo{year}{2017}\natexlab{}.
\newblock \showarticletitle{Practical Black-Box Attacks Against Machine
  Learning}. In \bibinfo{booktitle}{\emph{AsiaCCS}}. \bibinfo{publisher}{ACM},
  \bibinfo{pages}{506--519}.
\newblock


\bibitem[\protect\citeauthoryear{Papernot, McDaniel, Jha, Fredrikson, Celik,
  and Swami}{Papernot et~al\mbox{.}}{2016a}]%
        {PapernotMcDaniel2016-Limitations}
\bibfield{author}{\bibinfo{person}{Nicolas Papernot},
  \bibinfo{person}{Patrick~D. McDaniel}, \bibinfo{person}{Somesh Jha},
  \bibinfo{person}{Matt Fredrikson}, \bibinfo{person}{Z.~Berkay Celik}, {and}
  \bibinfo{person}{Ananthram Swami}.} \bibinfo{year}{2016}\natexlab{a}.
\newblock \showarticletitle{The Limitations of Deep Learning in Adversarial
  Settings}. In \bibinfo{booktitle}{\emph{EuroS\&P}}. \bibinfo{publisher}{IEEE
  Computer Society}, \bibinfo{pages}{372--387}.
\newblock


\bibitem[\protect\citeauthoryear{Papernot, McDaniel, Wu, Jha, and
  Swami}{Papernot et~al\mbox{.}}{2016b}]%
        {PapernotMcDaniel2016-Distillation}
\bibfield{author}{\bibinfo{person}{Nicolas Papernot},
  \bibinfo{person}{Patrick~D. McDaniel}, \bibinfo{person}{Xi Wu},
  \bibinfo{person}{Somesh Jha}, {and} \bibinfo{person}{Ananthram Swami}.}
  \bibinfo{year}{2016}\natexlab{b}.
\newblock \showarticletitle{Distillation as a Defense to Adversarial
  Perturbations Against Deep Neural Networks}. In
  \bibinfo{booktitle}{\emph{S\&P}}. \bibinfo{publisher}{IEEE Computer Society},
  \bibinfo{pages}{582--597}.
\newblock


\bibitem[\protect\citeauthoryear{Pedregosa, Varoquaux, Gramfort, Michel,
  Thirion, Grisel, Blondel, Prettenhofer, Weiss, Dubourg, VanderPlas, Passos,
  Cournapeau, Brucher, Perrot, and Duchesnay}{Pedregosa et~al\mbox{.}}{2011}]%
        {PedregosaVaroquaux2011}
\bibfield{author}{\bibinfo{person}{Fabian Pedregosa},
  \bibinfo{person}{Ga{\"{e}}l Varoquaux}, \bibinfo{person}{Alexandre Gramfort},
  \bibinfo{person}{Vincent Michel}, \bibinfo{person}{Bertrand Thirion},
  \bibinfo{person}{Olivier Grisel}, \bibinfo{person}{Mathieu Blondel},
  \bibinfo{person}{Peter Prettenhofer}, \bibinfo{person}{Ron Weiss},
  \bibinfo{person}{Vincent Dubourg}, \bibinfo{person}{Jake VanderPlas},
  \bibinfo{person}{Alexandre Passos}, \bibinfo{person}{David Cournapeau},
  \bibinfo{person}{Matthieu Brucher}, \bibinfo{person}{Matthieu Perrot}, {and}
  \bibinfo{person}{Edouard Duchesnay}.} \bibinfo{year}{2011}\natexlab{}.
\newblock \showarticletitle{{Scikit-learn}: {M}achine Learning in {Python}}.
\newblock \bibinfo{journal}{\emph{JMLR}}  \bibinfo{volume}{12}
  (\bibinfo{year}{2011}), \bibinfo{pages}{2825--2830}.
\newblock


\bibitem[\protect\citeauthoryear{Pei, Cao, Yang, and Jana}{Pei
  et~al\mbox{.}}{2017}]%
        {PeiCao2017}
\bibfield{author}{\bibinfo{person}{Kexin Pei}, \bibinfo{person}{Yinzhi Cao},
  \bibinfo{person}{Junfeng Yang}, {and} \bibinfo{person}{Suman Jana}.}
  \bibinfo{year}{2017}\natexlab{}.
\newblock \showarticletitle{{DeepXplore}: {A}utomated Whitebox Testing of Deep
  Learning Systems}. In \bibinfo{booktitle}{\emph{SOSP}}.
  \bibinfo{publisher}{ACM}, \bibinfo{pages}{1--18}.
\newblock


\bibitem[\protect\citeauthoryear{Quinlan}{Quinlan}{1986}]%
        {Quinlan1986}
\bibfield{author}{\bibinfo{person}{J.~Ross Quinlan}.}
  \bibinfo{year}{1986}\natexlab{}.
\newblock \showarticletitle{Induction of Decision Trees}.
\newblock \bibinfo{journal}{\emph{Machine Learning}}  \bibinfo{volume}{1}
  (\bibinfo{year}{1986}), \bibinfo{pages}{81--106}.
\newblock
Issue 1.


\bibitem[\protect\citeauthoryear{Roth, Kilcher, and Hofmann}{Roth
  et~al\mbox{.}}{2019}]%
        {RothKilcher2019}
\bibfield{author}{\bibinfo{person}{Kevin Roth}, \bibinfo{person}{Yannic
  Kilcher}, {and} \bibinfo{person}{Thomas Hofmann}.}
  \bibinfo{year}{2019}\natexlab{}.
\newblock \showarticletitle{The Odds Are Odd: {A} Statistical Test for
  Detecting Adversarial Examples}. In \bibinfo{booktitle}{\emph{ICML}}
  \emph{(\bibinfo{series}{PMLR})}, Vol.~\bibinfo{volume}{97}.
  \bibinfo{publisher}{PMLR}, \bibinfo{pages}{5498--5507}.
\newblock


\bibitem[\protect\citeauthoryear{Schapire}{Schapire}{1999}]%
        {Schapire1999}
\bibfield{author}{\bibinfo{person}{Robert~E. Schapire}.}
  \bibinfo{year}{1999}\natexlab{}.
\newblock \showarticletitle{A Brief Introduction to Boosting}. In
  \bibinfo{booktitle}{\emph{IJCAI}}. \bibinfo{publisher}{Morgan Kaufmann},
  \bibinfo{pages}{1401--1406}.
\newblock


\bibitem[\protect\citeauthoryear{Sen, Marinov, and Agha}{Sen
  et~al\mbox{.}}{2005}]%
        {SenMarinov2005}
\bibfield{author}{\bibinfo{person}{Koushik Sen}, \bibinfo{person}{Darko
  Marinov}, {and} \bibinfo{person}{Gul Agha}.} \bibinfo{year}{2005}\natexlab{}.
\newblock \showarticletitle{{CUTE}: {A} Concolic Unit Testing Engine for {C}}.
  In \bibinfo{booktitle}{\emph{ESEC/FSE}}. \bibinfo{publisher}{ACM},
  \bibinfo{pages}{263--272}.
\newblock


\bibitem[\protect\citeauthoryear{Sharma and Chen}{Sharma and Chen}{2018}]%
        {SharmaChen2018}
\bibfield{author}{\bibinfo{person}{Yash Sharma} {and} \bibinfo{person}{Pin{-}Yu
  Chen}.} \bibinfo{year}{2018}\natexlab{}.
\newblock \showarticletitle{Bypassing Feature Squeezing by Increasing Adversary
  Strength}.
\newblock \bibinfo{journal}{\emph{CoRR}}  \bibinfo{volume}{abs/1803.09868}
  (\bibinfo{year}{2018}).
\newblock


\bibitem[\protect\citeauthoryear{Siegmund, Siegmund, and Apel}{Siegmund
  et~al\mbox{.}}{2015}]%
        {SiegmundSiegmund2015}
\bibfield{author}{\bibinfo{person}{Janet Siegmund}, \bibinfo{person}{Norbert
  Siegmund}, {and} \bibinfo{person}{Sven Apel}.}
  \bibinfo{year}{2015}\natexlab{}.
\newblock \showarticletitle{Views on Internal and External Validity in
  Empirical Software Engineering}. In \bibinfo{booktitle}{\emph{ICSE}}.
  \bibinfo{publisher}{IEEE Computer Society}, \bibinfo{pages}{9--19}.
\newblock


\bibitem[\protect\citeauthoryear{Sun, Huang, and Kroening}{Sun
  et~al\mbox{.}}{2018a}]%
        {SunHuang2018}
\bibfield{author}{\bibinfo{person}{Youcheng Sun}, \bibinfo{person}{Xiaowei
  Huang}, {and} \bibinfo{person}{Daniel Kroening}.}
  \bibinfo{year}{2018}\natexlab{a}.
\newblock \showarticletitle{Testing Deep Neural Networks}.
\newblock \bibinfo{journal}{\emph{CoRR}}  \bibinfo{volume}{abs/1803.04792}
  (\bibinfo{year}{2018}).
\newblock


\bibitem[\protect\citeauthoryear{Sun, Wu, Ruan, Huang, Kwiatkowska, and
  Kroening}{Sun et~al\mbox{.}}{2018b}]%
        {SunWu2018}
\bibfield{author}{\bibinfo{person}{Youcheng Sun}, \bibinfo{person}{Min Wu},
  \bibinfo{person}{Wenjie Ruan}, \bibinfo{person}{Xiaowei Huang},
  \bibinfo{person}{Marta Kwiatkowska}, {and} \bibinfo{person}{Daniel
  Kroening}.} \bibinfo{year}{2018}\natexlab{b}.
\newblock \showarticletitle{Concolic Testing for Deep Neural Networks}. In
  \bibinfo{booktitle}{\emph{ASE}}. \bibinfo{publisher}{ACM},
  \bibinfo{pages}{109--119}.
\newblock


\bibitem[\protect\citeauthoryear{Szegedy, Vanhoucke, Ioffe, Shlens, and
  Wojna}{Szegedy et~al\mbox{.}}{2016}]%
        {SzegedyVanhoucke2016}
\bibfield{author}{\bibinfo{person}{Christian Szegedy}, \bibinfo{person}{Vincent
  Vanhoucke}, \bibinfo{person}{Sergey Ioffe}, \bibinfo{person}{Jonathon
  Shlens}, {and} \bibinfo{person}{Zbigniew Wojna}.}
  \bibinfo{year}{2016}\natexlab{}.
\newblock \showarticletitle{Rethinking the Inception Architecture for Computer
  Vision}. In \bibinfo{booktitle}{\emph{CVPR}}. \bibinfo{publisher}{IEEE
  Computer Society}, \bibinfo{pages}{2818--2826}.
\newblock


\bibitem[\protect\citeauthoryear{Szegedy, Zaremba, Sutskever, Bruna, Erhan,
  Goodfellow, and Fergus}{Szegedy et~al\mbox{.}}{2014}]%
        {SzegedyZaremba2014}
\bibfield{author}{\bibinfo{person}{Christian Szegedy},
  \bibinfo{person}{Wojciech Zaremba}, \bibinfo{person}{Ilya Sutskever},
  \bibinfo{person}{Joan Bruna}, \bibinfo{person}{Dumitru Erhan},
  \bibinfo{person}{Ian~J. Goodfellow}, {and} \bibinfo{person}{Rob Fergus}.}
  \bibinfo{year}{2014}\natexlab{}.
\newblock \showarticletitle{Intriguing Properties of Neural Networks}. In
  \bibinfo{booktitle}{\emph{ICLR}}.
\newblock


\bibitem[\protect\citeauthoryear{Tian, Pei, Jana, and Ray}{Tian
  et~al\mbox{.}}{2018}]%
        {TianPei2018}
\bibfield{author}{\bibinfo{person}{Yuchi Tian}, \bibinfo{person}{Kexin Pei},
  \bibinfo{person}{Suman Jana}, {and} \bibinfo{person}{Baishakhi Ray}.}
  \bibinfo{year}{2018}\natexlab{}.
\newblock \showarticletitle{{DeepTest}: {A}utomated Testing of
  Deep-Neural-Network-Driven Autonomous Cars}. In
  \bibinfo{booktitle}{\emph{ICSE}}. \bibinfo{publisher}{ACM},
  \bibinfo{pages}{303--314}.
\newblock


\bibitem[\protect\citeauthoryear{Wang, Dong, Sun, Wang, and Zhang}{Wang
  et~al\mbox{.}}{2019}]%
        {WangDong2019}
\bibfield{author}{\bibinfo{person}{Jingyi Wang}, \bibinfo{person}{Guoliang
  Dong}, \bibinfo{person}{Jun Sun}, \bibinfo{person}{Xinyu Wang}, {and}
  \bibinfo{person}{Peixin Zhang}.} \bibinfo{year}{2019}\natexlab{}.
\newblock \showarticletitle{Adversarial Sample Detection for Deep Neural
  Network Through Model Mutation Testing}. In \bibinfo{booktitle}{\emph{ICSE}}.
  \bibinfo{publisher}{IEEE Computer Society/ACM}, \bibinfo{pages}{1245--1256}.
\newblock


\bibitem[\protect\citeauthoryear{Xie, Ma, Juefei{-}Xu, Xue, Chen, Liu, Zhao,
  Li, Yin, and See}{Xie et~al\mbox{.}}{2019}]%
        {XieMa2019}
\bibfield{author}{\bibinfo{person}{Xiaofei Xie}, \bibinfo{person}{Lei Ma},
  \bibinfo{person}{Felix Juefei{-}Xu}, \bibinfo{person}{Minhui Xue},
  \bibinfo{person}{Hongxu Chen}, \bibinfo{person}{Yang Liu},
  \bibinfo{person}{Jianjun Zhao}, \bibinfo{person}{Bo Li},
  \bibinfo{person}{Jianxiong Yin}, {and} \bibinfo{person}{Simon See}.}
  \bibinfo{year}{2019}\natexlab{}.
\newblock \showarticletitle{{DeepHunter}: {A} Coverage-Guided Fuzz Testing
  Framework for Deep Neural Networks}. In \bibinfo{booktitle}{\emph{ISSTA}}.
  \bibinfo{publisher}{ACM}, \bibinfo{pages}{146--157}.
\newblock


\bibitem[\protect\citeauthoryear{Xu, Evans, and Qi}{Xu et~al\mbox{.}}{2018}]%
        {XuEvans2018}
\bibfield{author}{\bibinfo{person}{Weilin Xu}, \bibinfo{person}{David Evans},
  {and} \bibinfo{person}{Yanjun Qi}.} \bibinfo{year}{2018}\natexlab{}.
\newblock \showarticletitle{Feature Squeezing: {D}etecting Adversarial Examples
  in Deep Neural Networks}. In \bibinfo{booktitle}{\emph{NDSS}}.
  \bibinfo{publisher}{The Internet Society}.
\newblock


\bibitem[\protect\citeauthoryear{Zantedeschi, Nicolae, and Rawat}{Zantedeschi
  et~al\mbox{.}}{2017}]%
        {ZantedeschiNicolae2017}
\bibfield{author}{\bibinfo{person}{Valentina Zantedeschi},
  \bibinfo{person}{Maria{-}Irina Nicolae}, {and} \bibinfo{person}{Ambrish
  Rawat}.} \bibinfo{year}{2017}\natexlab{}.
\newblock \showarticletitle{Efficient Defenses Against Adversarial Attacks}. In
  \bibinfo{booktitle}{\emph{AISec@CCS}}. \bibinfo{publisher}{ACM},
  \bibinfo{pages}{39--49}.
\newblock


\bibitem[\protect\citeauthoryear{Zhang, Zhang, Zhang, Liu, and Khurshid}{Zhang
  et~al\mbox{.}}{2018}]%
        {ZhangZhang2018}
\bibfield{author}{\bibinfo{person}{Mengshi Zhang}, \bibinfo{person}{Yuqun
  Zhang}, \bibinfo{person}{Lingming Zhang}, \bibinfo{person}{Cong Liu}, {and}
  \bibinfo{person}{Sarfraz Khurshid}.} \bibinfo{year}{2018}\natexlab{}.
\newblock \showarticletitle{{DeepRoad}: {GAN}-Based Metamorphic Testing and
  Input Validation Framework for Autonomous Driving Systems}. In
  \bibinfo{booktitle}{\emph{ASE}}. \bibinfo{publisher}{ACM},
  \bibinfo{pages}{132--142}.
\newblock


\end{thebibliography}


\end{document}